\documentclass{article}

\usepackage[T1]{fontenc}
\usepackage[utf8]{inputenc}
\usepackage[english]{babel}  
\usepackage{microtype}       
\usepackage{newunicodechar}
\usepackage{xcolor}

\usepackage{graphicx}
\usepackage{subcaption}      
\usepackage{booktabs}        
\usepackage{stfloats}        

\usepackage{nicefrac}        
\usepackage{bm}              
\usepackage{amsmath}         
\usepackage{amssymb}
\usepackage{xspace}
\usepackage{siunitx}        

\usepackage{varioref}
\usepackage{hyperref}
\usepackage[nameinlink]{cleveref}
\usepackage{url}         

\addto\extrasenglish{

}

\crefname{chapter}{Chapter}{Chapters}
\crefname{section}{Section}{Sections}
\crefname{subsection}{Section}{Sections}
\crefname{subsubsection}{Section}{Sections}
\crefname{figure}{Figure}{Figures}
\crefname{table}{Table}{Tables}
\crefname{subfigure}{Figure}{Figures}
\crefname{page}{Page}{Pages}
\crefname{equation}{Equation}{Equations}
\crefname{appendix}{Appendix}{Appendix}

\newlength{\textfloatsepsave} 

\DeclareMathOperator*{\softmax}{\text{softmax}}

\newcommand{\p}{p}                      
\newcommand{\x}{\ensuremath{\mathbf{x}}\xspace}
\newcommand{\z}{\ensuremath{\mathbf{z}}\xspace}
\newcommand{\m}{\ensuremath{\mathbf{m}}\xspace}

\newcommand{\textOrMath}[2]{{\ifmmode{#1}\else{#2}\fi}}

\newcommand{\modelname}{IODINE\xspace}
\newcommand{\modellongname}{Iterative Object Decomposition Inference NEtwork}

\newcommand{\kgets}{\overset{\scriptscriptstyle k}{\bm{\gets}}}
\newcommand{\ksim}{\overset{\scriptscriptstyle k}{\bm{\sim}}}

\newunicodechar{ℕ}{\ensuremath{\mathbb{N}}}
\newunicodechar{ℤ}{\ensuremath{\mathbb{Z}}}
\newunicodechar{ℝ}{\ensuremath{\mathbb{R}}}
\newunicodechar{ℂ}{\ensuremath{\mathbb{C}}}

\newunicodechar{⊂}{\subset}
\newunicodechar{∈}{\in}
\newunicodechar{∪}{\cup}
\newunicodechar{∩}{\cap}
\newunicodechar{∇}{\nabla}

\newunicodechar{α}{\alpha}
\newunicodechar{β}{\beta}
\newunicodechar{γ}{\gamma}
\newunicodechar{δ}{\delta}
\newunicodechar{ε}{\epsilon}
\newunicodechar{ζ}{\zeta}
\newunicodechar{η}{\eta}
\newunicodechar{θ}{\theta}
\newunicodechar{ι}{\iota}
\newunicodechar{κ}{\kappa}
\newunicodechar{λ}{\lambda}
\newunicodechar{μ}{\mu}
\newunicodechar{ν}{\nu}
\newunicodechar{ο}{o}
\newunicodechar{π}{\pi}
\newunicodechar{ρ}{\rho}
\newunicodechar{σ}{\sigma}
\newunicodechar{τ}{\tau}
\newunicodechar{υ}{\upsilon}
\newunicodechar{φ}{\varphi}
\newunicodechar{χ}{\chi}
\newunicodechar{ψ}{\psi}
\newunicodechar{ω}{\omega}
\newunicodechar{ξ}{\xi}
\newunicodechar{ς}{\varsigma}
\newunicodechar{Ω}{\Omega}
\newunicodechar{Ξ}{\Xi}
\newunicodechar{Π}{\Pi}
\newunicodechar{Ψ}{\Psi}
\newunicodechar{Γ}{\Gamma}
\newunicodechar{Φ}{\Phi}
\newunicodechar{Σ}{\Sigma}
\newunicodechar{Δ}{\Delta}
\newunicodechar{Θ}{\Theta}
\newunicodechar{Λ}{\Lambda}

\usepackage[accepted]{icml2019}

\icmltitlerunning{Variational Iterative Multi-Object Representation Learning}

\begin{document}

\twocolumn[
\icmltitle{Multi-Object Representation Learning with Iterative Variational Inference}

\icmlsetsymbol{equal}{*}

\begin{icmlauthorlist}
\icmlauthor{Klaus Greff}{idsia,atdm}
\icmlauthor{Raphaël Lopez Kaufman}{dm}
\icmlauthor{Rishabh Kabra}{dm}
\icmlauthor{Nick Watters}{dm}
\icmlauthor{Chris Burgess}{dm}
\icmlauthor{Daniel Zoran}{dm}
\icmlauthor{Loic Matthey}{dm}
\icmlauthor{Matthew Botvinick}{dm}
\icmlauthor{Alexander Lerchner}{dm}
\end{icmlauthorlist}

\icmlaffiliation{idsia}{The Swiss AI lab IDSIA, Lugano, Switzerland}
\icmlaffiliation{dm}{DeepMind, London, UK}
\icmlaffiliation{atdm}{Work done at DeepMind}

\icmlcorrespondingauthor{Klaus Greff}{klaus.greff@startmail.com}

\icmlkeywords{Machine Learning, Multi-Object, Variational Inference, Perceptual Grouping, Representation Learning, Disentanglement, Unsupervised Learning}

\vskip 0.3in
]

\printAffiliationsAndNotice{}  

\begin{abstract}
Human perception is structured around objects which form the basis for our higher-level cognition and impressive systematic generalization abilities.
Yet most work on representation learning focuses on feature learning without even considering multiple objects, or treats segmentation as an (often supervised) preprocessing step.
Instead, we argue for the importance of learning to segment and represent objects \emph{jointly}.
We demonstrate that, starting from the simple assumption that a scene is composed of multiple entities, it is possible to learn to segment images into interpretable objects with disentangled representations.
Our method learns -- without supervision -- to inpaint occluded parts, and extrapolates to scenes with more objects and to unseen objects with novel feature combinations. 
We also show that, due to the use of iterative variational inference, our system is able to learn multi-modal posteriors for ambiguous inputs and extends naturally to sequences.
\end{abstract}

\section{Introduction}
\label{sec:introduction}

\begin{figure}[t]
    \centering
    \includegraphics[width=0.85\columnwidth]{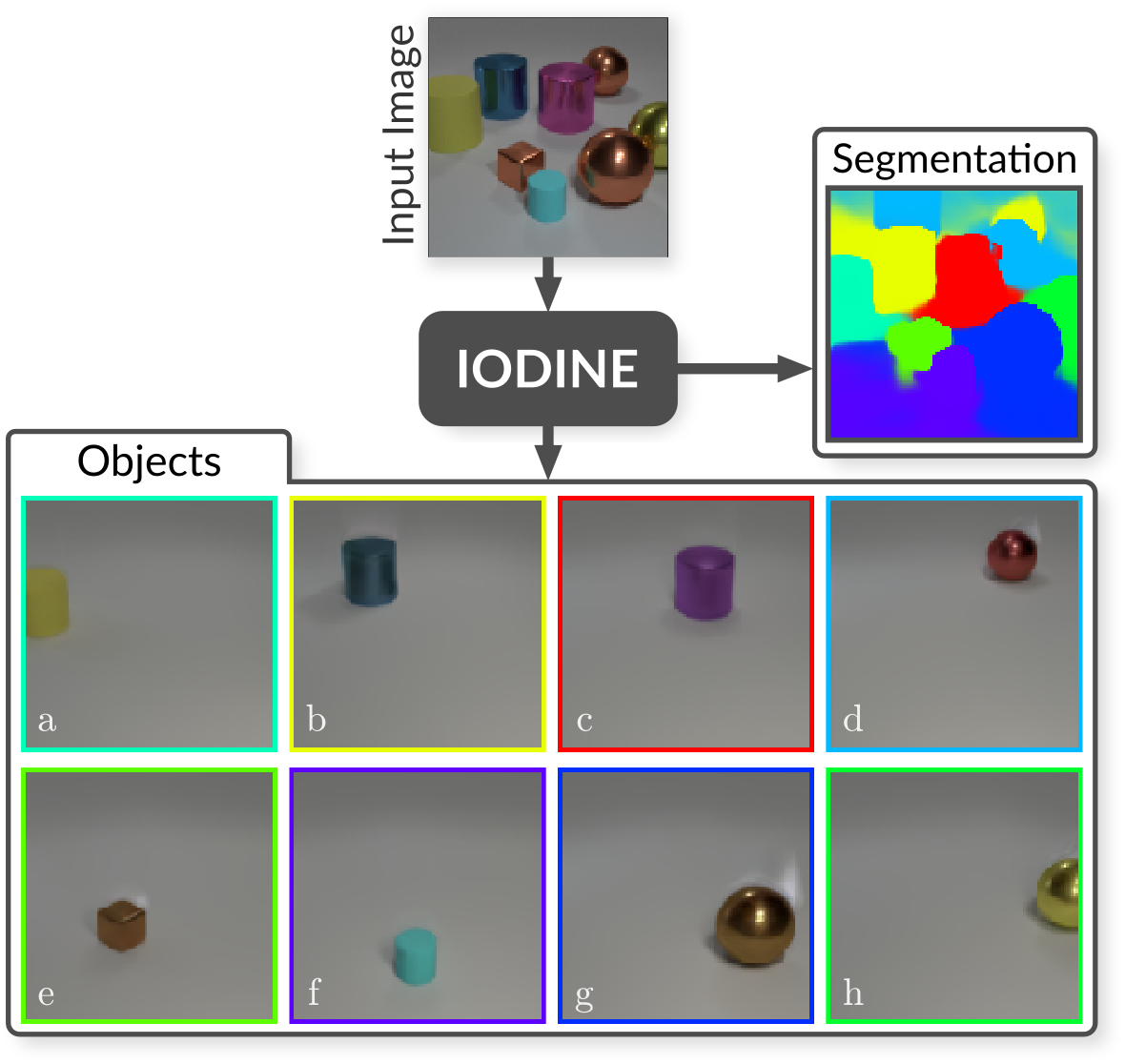}
    \vspace*{-7pt}
    \caption{Object decomposition of an image from the CLEVR dataset by \modelname. The model is able to decompose the image into separate objects in an unsupervised manner, inpainting occluded objects in the process (see slots (d), (e) and (h)).}
    \label{fig:preview}
    \vspace*{-15pt}
\end{figure}

Learning good representations of complex visual scenes is a challenging problem for artificial intelligence that is far from solved. 
Recent breakthroughs in unsupervised representation learning \cite{higgins2017beta, makhzani2015adversarial,chen2016infogan} tend to focus on data where a single object of interest is placed in front of some background (e.g. dSprites, 3D Chairs, CelebA). 
Yet in general, visual scenes contain a variable number of objects arranged in various spatial configurations, and often with partial occlusions (e.g., CLEVR, \citealt{johnson2017clevr}; see \cref{fig:preview}). 
This motivates the question: what forms a good representation of a scene with multiple objects? 
In line with recent advances \cite{burgess2019monet, steenkiste2018relational, eslami2016attend}, we maintain that discovery of objects in a scene should be considered a crucial aspect of representation learning, rather than treated as a separate problem.

We approach the problem from a spatial mixture model perspective \cite{greff2017neural} and use amortized iterative refinement \cite{marino2018iterative} of latent object representations within a variational framework \cite{rezende2014stochastic, kingma2013auto}.
We encode our basic intuition about the existence of objects into the structure of our model, which simultaneously facilitates their discovery and efficient representation in a fully data-driven, unsupervised manner.
We name the resulting architecture \textbf{\modelname} (short for \modellongname). 

\modelname can segment complex scenes and learn disentangled object features without supervision on datasets like CLEVR, Objects Room \cite{burgess2019monet}, and Tetris (see \cref{appendix:data}).
We show systematic generalization to more objects than included in the training regime, as well as objects formed with unseen feature combinations. 
This highlights the benefits of multi-object representation learning by comparison to a VAE's single-slot representations. 
We also justify how the sampling used in iterative refinement lends to resolving multi-modal and multi-stable decomposition.

\section{Method}
\label{sec:method}
We first express multi-object representation learning within the framework of generative modelling (\cref{sec:method:desiderata}).
Then, building upon the successful Variational AutoEncoder framework (VAEs; \citealt{rezende2014stochastic, kingma2013auto}), we leverage variational inference to jointly learn both the generative and inference model (\cref{sec:method:inference}).
There we also discuss the particular challenges that arise for inference in a multi-object context and show how they can be solved using iterative amortization.
Finally, in \cref{sec:method:training} we bring all elements together and show how the complete system can be trained end-to-end.

\subsection{Multi-Object Representations}
\label{sec:method:desiderata}
Flat vector representations as used by standard VAEs are inadequate for capturing the combinatorial object structure that many datasets exhibit. To achieve the kind of systematic generalization that is so natural for humans, we propose employing a \emph{multi-slot} representation where each slot shares the underlying representation format, and each would ideally describe an independent part of the input.
Consider the example in \cref{fig:preview}: by construction, the scene consists of 8 objects, each with its own properties such as shape, size, position, color and material. 
To split objects, a flat representation would have to represent each object using separate feature dimensions.
But this neglects the simple and (to us) trivial fact that they are interchangeable objects with common properties.

\begin{figure}[t]
    \begin{subfigure}[b]{0.15\linewidth}
        \centering
        \includegraphics[height=2cm]{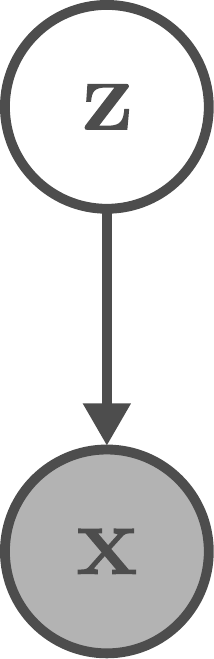}
        \caption{VAE}
        \label{fig:pgm_vae}
    \end{subfigure}
    \hfill
    \begin{subfigure}[b]{0.35\linewidth}
        \centering
        \includegraphics[height=2cm]{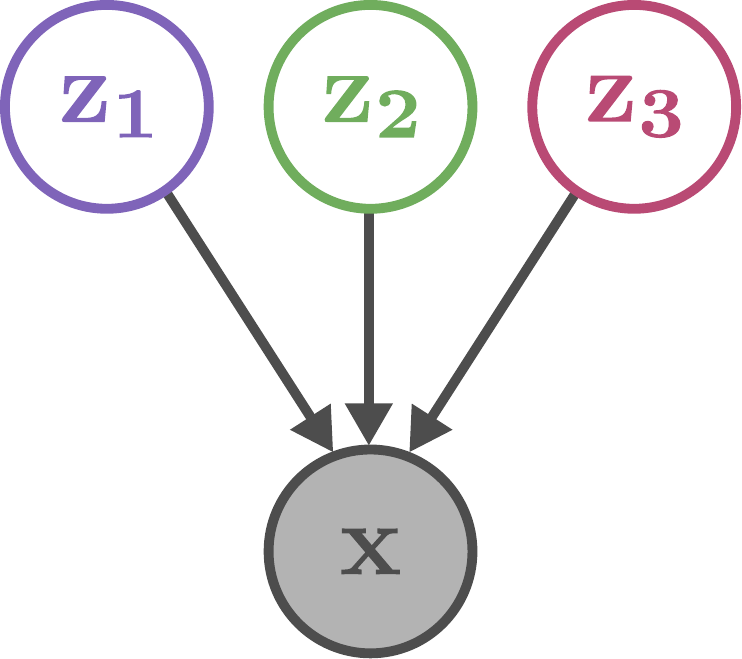}
        \caption{Multi-object VAE}
        \label{fig:pgm_iasmvae}
    \end{subfigure}
    \hfill
    \begin{subfigure}[b]{0.4\linewidth}
        \centering
        \includegraphics[height=2cm]{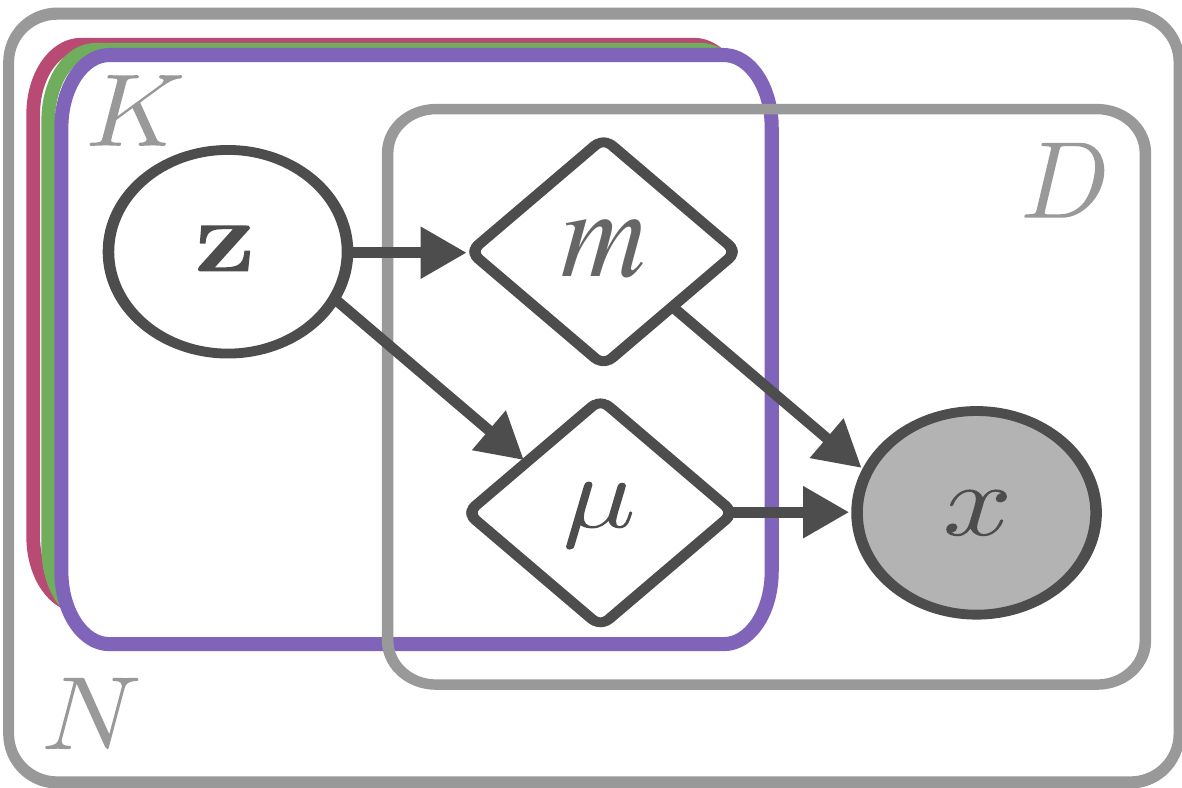}
        \caption{\modelname}
        \label{fig:pgm_detail}
    \end{subfigure}
    \begin{subfigure}[b]{0.95\linewidth}
        \centering
        \includegraphics[width=\linewidth]{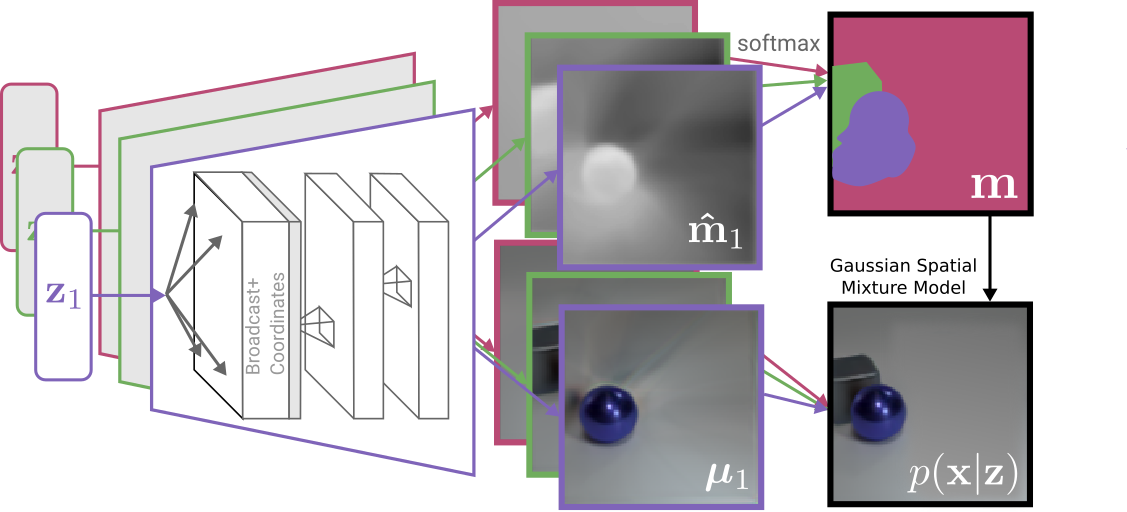}
        \caption{\modelname neural architecture.}
        \label{fig:decoder}
    \end{subfigure}
    \vspace*{-8pt}
    \caption{Generative model illustrations. (a) A regular VAE decoder. (b) A hypothetical multi-object VAE decoder that recomposes the scene from three objects. (c) \modelname's multi-object decoder showing latent vectors (denoted $\bm{z}$) corresponding to $K$ objects refined over $N$ iterations from images of dimension $D$. The deterministic pixel-wise means and masks are denoted $\bm{μ}$ and $\m$ respectively. 
    (d) The neural architecture of the \modelname's multi-object spatial mixture decoder.}
    \label{fig:generativeModel}
    \vspace{-10pt}
\end{figure}

\paragraph{Generative Model}
We represent each scene with $K$ latent object representations $\z_k ∈ ℝ^M$ that collaborate to generate the input image $\x ∈ ℝ^D$ (c.f. \cref{fig:pgm_iasmvae}).
The $\z_k$ are assumed to be independent and their generative mechanism is shared such that any ordering of them produces the same image (i.e. entailing permutation invariance).
Objects distinguished in this way can easily be compared, reused and recombined, thus facilitating combinatorial generalization.

The image $\x$ is modeled with a spatial Gaussian mixture model where each mixing component (slot) corresponds to a single object.
That means each object vector $\z_k$ is decoded into a pixel-wise mean $\mu_{ik}$ (the appearance of the object) and a pixel-wise assignment $m_{ik} = \p(C=k|\z_k)$ (the segmentation mask; c.f. \cref{fig:pgm_detail}). 
Assuming that the pixels $i$ are independent conditioned on $\z$, the likelihood thus becomes:
\begin{equation}
    \p(\x | \z) = \prod_{i=1}^{D}\sum_{k=1}^{K} m_{ik} \mathcal{N}(x_i; μ_{ik}, σ^2),
    \label{eq:likelihood}
\end{equation}
where we use a global fixed variance $σ^2$ for all pixels.

\paragraph{Decoder Structure}
Our decoder network structure directly reflects the structure of the generative model. See \cref{fig:decoder} for an illustration.
Each object latent $\z_k$ is decoded separately into pixel-wise means $\bm{\mu}_k$ and mask-logits $\mathbf{\hat{m}}_k$, which we then normalize using a softmax operation applied across slots such that the masks $\mathbf{m}_k$ for each pixel sum to 1.
Together, $\bm{\mu}$ and $\mathbf{m}$ parameterize the spatial mixture distribution as defined in \cref{eq:likelihood}.
For the network architecture we use a broadcast decoder~\cite{watters2019}, which spatially replicates the latent vector $\z_k$, appends two coordinate channels (ranging from $-1$ to $1$ horizontally and vertically), and applies a series of size-preserving convolutional layers.
This structure encourages disentangling the position across the image from other features such as color or texture, and generally supports disentangling.
All slots $k$ share weights to ensure a common format, and are independently decoded, up until the mask normalization. 

\subsection{Inference}
\label{sec:method:inference}
Similar to VAEs, we use amortized variational inference to get an approximate posterior $q_\lambda(\z|\x)$ parameterized as a Gaussian with parameters $\bm{λ} = \{\bm{μ_z}, \bm{σ_z}\}$.
However, our object-oriented generative model poses a few specific challenges for the inference process:
Firstly, being a (spatial) mixture model, we need to infer both the components (i.e. object appearance) and the mixing (i.e. object segmentation).
This type of problem is well known, for example in clustering and image segmentation, and is traditionally tackled as an iterative procedure, because there are no efficient direct solutions.
A related second problem is that any slot can, in principle, explain any pixel.
Once a pixel is explained by one of the slots, the others don't need to account for it anymore.
This explaining-away property complicates the inference by strongly coupling it across the individual slots.
Finally, slot permutation invariance induces a multimodal posterior with at least one mode per slot permutation. 
This is problematic, since our approximate posterior $q_\lambda(\z|\x)$ is parameterized as a unimodal distribution.
For all the above reasons, the standard feed-forward VAE inference model is inadequate for our case, so we consider a more powerful method for inference.

\paragraph{Iterative Inference}

\begin{figure}
    \centering
    \includegraphics[width=\columnwidth]{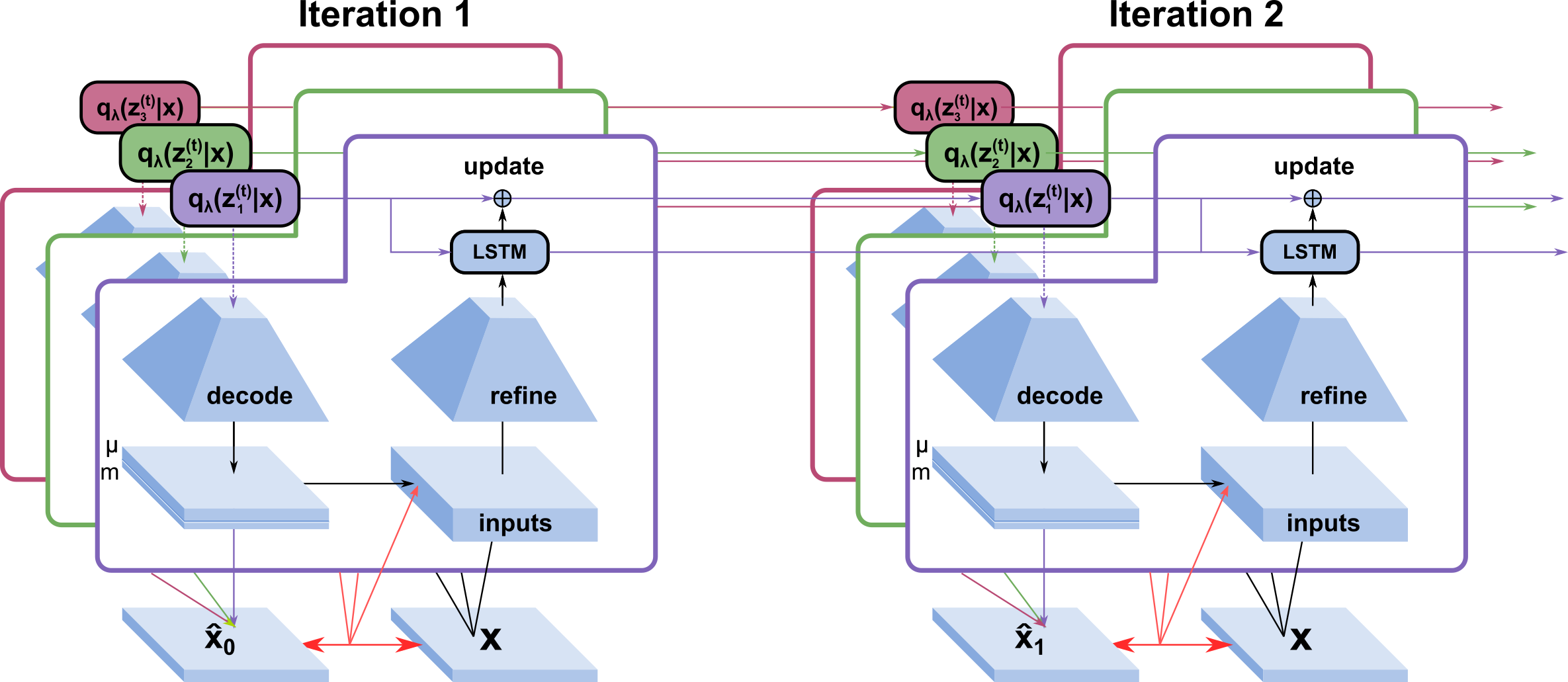}
    \vspace*{-5pt}
    \caption{Illustration of the iterative inference procedure.}
    \label{fig:iterations_illustration}
    \vspace*{-10pt}
\end{figure}

The basic idea of iterative inference is to start with an arbitrary guess for the posterior parameters $\bm{λ}$, and then iteratively refine them using the input and samples from the current posterior estimate.
We build on the framework of iterative amortized inference~\cite{marino2018iterative}, which uses a trained refinement network $f_\phi$.
Unlike Marino et al., we consider only additive updates to the posterior and use several salient auxiliary inputs $\mathbf{a}$ to the refinement network (instead of just $\nabla_\lambda\mathcal{L}$).
We update the posterior of the $K$ slots independently and in parallel (indicated by  $\kgets$ and $\ksim$), as follows:
\begin{align}
   \z^{(t)}_k  &\ksim q_λ(\z^{(t)}_k | \x)\label{eq:sample}\\ 
   \bm{λ}^{(t+1)}_k &\kgets \bm{λ}^{(t)}_k + f_\phi(\z^{(t)}_k, \x, \mathbf{a}_k)\,,  \label{eq:update}
\end{align}
Thus the only place where the slots interact are at the input level.
As refinement network $f_\phi$ we use a convolutional network followed by an LSTM (see \cref{appendix:model} for details).
Instead of amortizing the posterior directly (as in a regular VAE encoder), the refinement network can be thought of as amortizing the gradient of the posterior~\cite{marino2018general}.
The alternating updates to $q_\lambda(\z|\x)$ and $p(\x|\z)$ are also akin to message passing. 

\setlength{\textfloatsep}{0pt}

\begin{algorithm}
\begin{algorithmic}
\renewcommand{\COMMENT}[2][.25\linewidth]{%
  \leavevmode\hfill\makebox[#1][l]{\textcolor{gray}{//~#2}}}
    \STATE \textbf{Input:} image $\x$, hyperparamters $K$, $T$, $\sigma^2$
    \STATE \textbf{Input:} trainable parameters $\bm{λ}^{(1)}$, $\bm{θ}$, $\bm{\phi}$
    \STATE \textbf{Initialize:} $\mathbf{h}^{(1)}_k \kgets \mathbf{0}$
    \FOR{$t=1$ {\bfseries to} $T$}
       \STATE $\z^{(t)}_k \ksim q_λ(\z_k^{(t)}|\x)$ \COMMENT{Sample}
       \STATE $\bm{μ}_k^{(t)}, \bm{\hat{m}}_k^{(t)} \kgets g_θ(\z^{(t)}_k)$ \COMMENT{Decode}
       \STATE $\bm{m}^{(t)} \gets \softmax_k(\bm{\hat{m}}^{(t)}_k)$  \COMMENT{Masks}
       \STATE $p(\x|\z^{(t)}) \gets \sum_k \bm{m}^{(t)}_k \mathcal{N}(\x; \bm{μ}^{(t)}_k, σ^2)$ \COMMENT{Likelihood}
       \STATE $\mathcal{L}^{(t)} \gets D_{KL}(q_λ(\z^{(t)}|\x) || p(\z)) - \log p(\x|\z^{(t)})$
       
       \STATE $\mathbf{a_k} \kgets \text{inputs}(\x, \z_k^{(t)}, \bm{λ}_k^{(t)})$ \COMMENT{Inputs}
       \STATE $\bm{λ}^{(t+1)}_k, \mathbf{h}^{(t+1)} \kgets  f_\phi(\mathbf{a}_k, \mathbf{h}_k^{(t)})$  \COMMENT{Refinement}
    \ENDFOR
\end{algorithmic}
\caption{\modelname Pseudocode.}\label{alg:iasmvae}
\end{algorithm}
\setlength{\textfloatsep}{\textfloatsepsave}

\paragraph{Inputs}
\label{sec:method:inputs}
For each slot $k$ we feed a set of auxiliary inputs $\bm{a}_k$ to the refinement network $f_\phi$ which then computes an update for the posterior $\bm{\lambda}_k$.
Crucially, we include gradient information about the ELBO in the inputs, as it conveys information about what is not yet explained by other slots. 
Omitting the superscript $(t)$ for clarity, the auxiliary inputs $\mathbf{a}_k$ are (see \cref{appendix:model} for details):
image $\x$,  means $\bm{μ}_k$, masks $\m_k$, mask-logits $\bm{\hat{m}}_k$, mean gradient $\nabla_{\bm{μ}_k}\mathcal{L}$, mask gradient $\nabla_{\m_k}\mathcal{L}$, posterior gradient $\nabla_{\bm{λ}_k}\mathcal{L}$, posterior mask $\p(\m_k|\x,\bm{μ}) = \frac{\p(x|μ_k)}{\sum_j \p(x|μ_j)}$, pixelwise likelihood $\p(\x|\z)$, leave-one-out likelihood $\p(\x|\z_{i \neq k})$, and two coordinate channels like in the decoder.

With the exception of $\nabla_{\bm{λ}_k}\mathcal{L}$, these are all image-sized and cheap to compute, so we feed them as additional input-channels into the refinement network.
The approximate gradient $\nabla_{\bm{λ}_k}\mathcal{L}$ is computed using the reparameterization trick by a backward pass through the generator network.
This is computationally quite expensive, but we found that this information helps to significantly improve training of the refinement network.
This input is the same size as the posterior $\bm{\lambda}_k$ and is fed to the LSTM part of the refinement network.
Like \citet{marino2018iterative} we found it beneficial to normalize the gradient-based inputs with LayerNorm~\cite{ba2016layer}.
See \cref{sec:results:robustness} for an ablation study.

\subsection{Training}
\label{sec:method:training}
We train the parameters of the decoder ($\bm{θ}$), of the refinement network ($\bm{\phi}$), and of the initial posterior ($\bm{λ}^{(1)}$) by gradient descent through the unrolled iterations. 
In principle, it is enough to minimize the final negative ELBO $\mathcal{L}^{T}$, but we found it beneficial to use a weighted sum which also includes earlier terms:
\vspace{-5pt}
\begin{equation}
    \mathcal{L}_{\text{total}} = \sum_{t=1}^T \frac{t}{T} \mathcal{L}^{(t)}.
\end{equation}
Each refinement step of \modelname uses gradient information to optimize the posterior $\bm{\lambda}$.
Unfortunately, backpropagating through this process leads to numerical instabilities connected to double derivatives like $∇_Θ∇_z\mathcal{L}$. 
We found that this problem can be mitigated by dropping the double derivative terms, i.e. stopping the gradients from backpropagating through the gradient-inputs $\nabla_{\bm{μ}_k}\mathcal{L}$, $\nabla_{\m_k}\mathcal{L}$, and $\nabla_{\bm{λ}_k}\mathcal{L}$ (see \cref{appendix:model} for details).

\section{Related Work}
\label{sec:related}
Representation learning \cite{bengio2013representation} has received much attention and has seen several recent breakthroughs. This includes disentangled representations through the use of $β$-VAEs~\cite{higgins2017beta}, adversarial autoencoders ~\cite{makhzani2015adversarial}, Factor VAEs~\cite{kim2018disentangling}, and improved generalization through non-euclidean embeddings \cite{nickel2017poincare}.
However, most advances have focused on the feature-level structure of representations, and do not address the issue of representing multiple, potentially repeating objects, which we tackle here.

Another line of work is concerned with obtaining segmentations of images, usually without considering representation learning.
This has led to impressive results on real-world images, however, many approaches (such as ``semantic segmentation'' or object detection) rely on supervised signals \cite{Girshick2015, He2017, Redmon2018}, while others require hand-engineered features \cite{shi2000normalized, felzenszwalb2004efficient}.
In contrast, as we learn to both segment and represent, our method can perform inpainting (\cref{fig:preview}) and deal with ambiguity (\cref{fig:tetris}), going beyond what most methods relying on feature engineering are currently able to do.

Works tackling the full problem of scene representation are rarer.
Probabilistic programming based approaches, like stroke-based character generation \cite{lake2015human} or 3D indoor scene rendering \cite{pero2012bayesian}, have produced appealing results, but require carefully engineered generative models, which are typically not fully learned from data.
Work on end-to-end models has shown promise in using autoregressive inference or generative approaches \cite{eslami2016attend,gregor2015draw}, including the recent MONet~\cite{burgess2019monet}.
Few methods can achieve similar comparable with the complexity of the scenes we consider here, apart from MONet.
\cref{sec:results:segmentation} shows a preliminary comparison between MONet and \modelname, and we discuss their relationship further in \cref{appendix:discussion:monet}.

Two other methods related to ours are Neural Expectation Maximization \cite{greff2017neural} (along with its sequential and relational extensions \cite{steenkiste2018relational}) and Tagger~\cite{greff2016tagger}. NEM uses recurrent neural networks to amortize expectation maximization for a spatial mixture model. However, NEM variants fail to cope with colored scenes, as we note in our comparison in \cref{sec:results:segmentation}. Tagger also uses iterative inference to segment and represent images based on a denoising training objective. 
We disregard Tagger for our comparison, because (1) its use of a Ladder network means that there is no bottleneck and thus no explicit object representations, and (2) without adapting it to a convolutional architecture, it does not scale to larger images (Tagger would require $\approx600\text{M}$ weights for CLEVR).

\section{Results}
\label{sec:results}

\begin{figure}
    \centering
    \includegraphics[width=\linewidth]{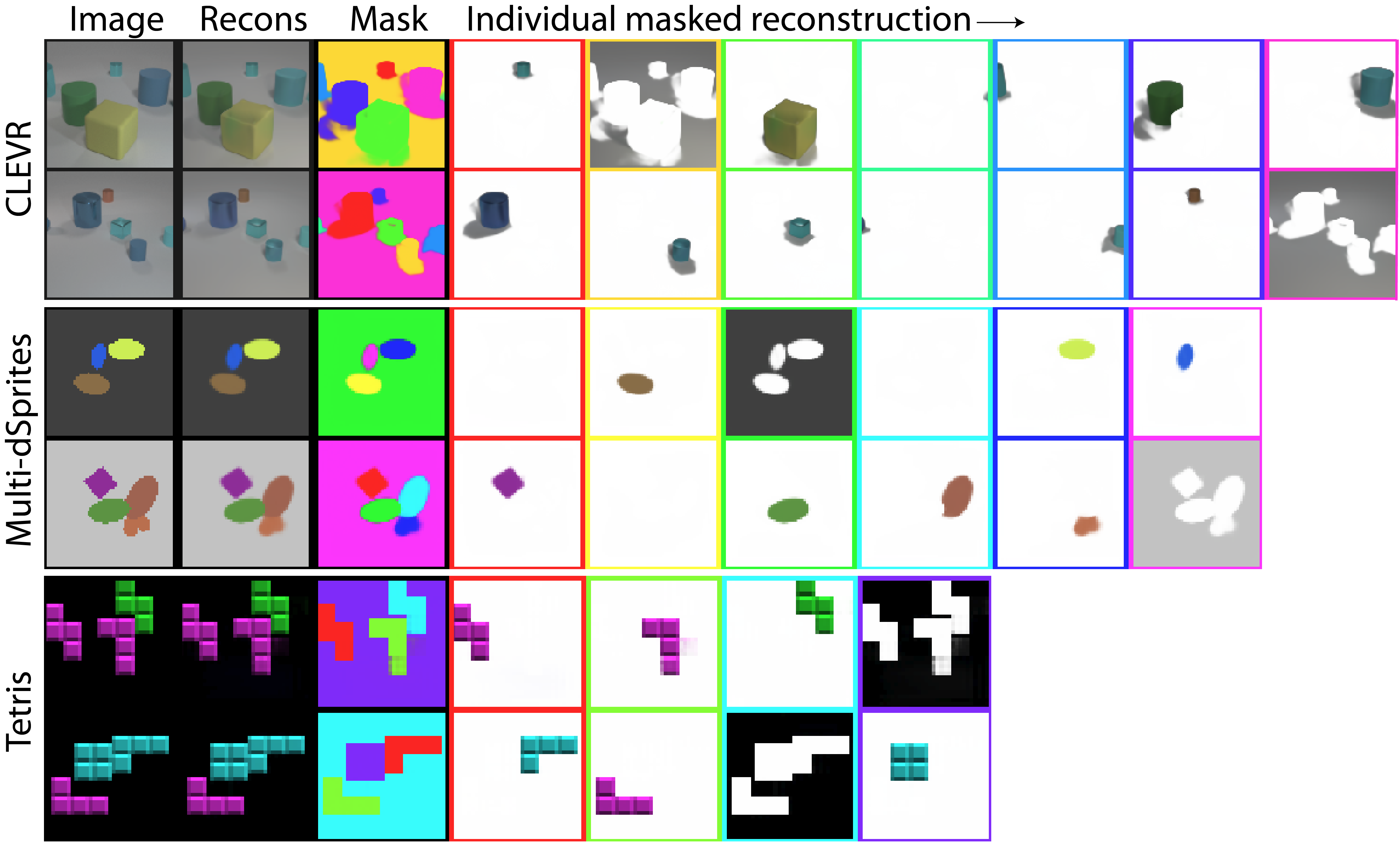}
    \vspace*{-20pt}
    \caption{\modelname segmentations and object reconstructions on CLEVR6 (top), Multi-dSprites (middle), and Tetris (bottom). The individual masked reconstruction slots represent objects separately (along with their shadow on CLEVR). Border colours are matched to the segmentation mask on the left.}
    \vspace*{-5pt}
    \label{fig:decompositions}
\end{figure}

\begin{figure}
    \centering
    \includegraphics[width=0.9\columnwidth]{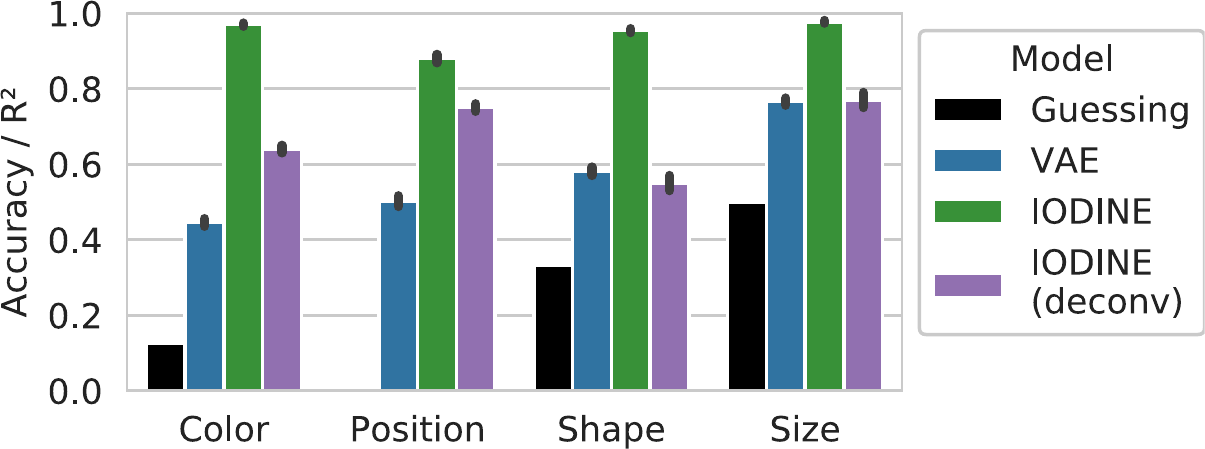}
    \vspace{-8pt}
    \caption{Prediction accuracy / $R^2$ score for the factor regression on CLEVR6. Position is continuous; the rest are categorical with 8 colors, 3 shapes, and 2 sizes. IODINE (deconv) does not use spatial broadcasting in the decoder (see \cref{sec:results:ablation}).}
    \label{fig:factors}
    \vspace*{-12pt}
\end{figure}

\begin{figure*}
    \centering
    \includegraphics[width=0.99\linewidth]{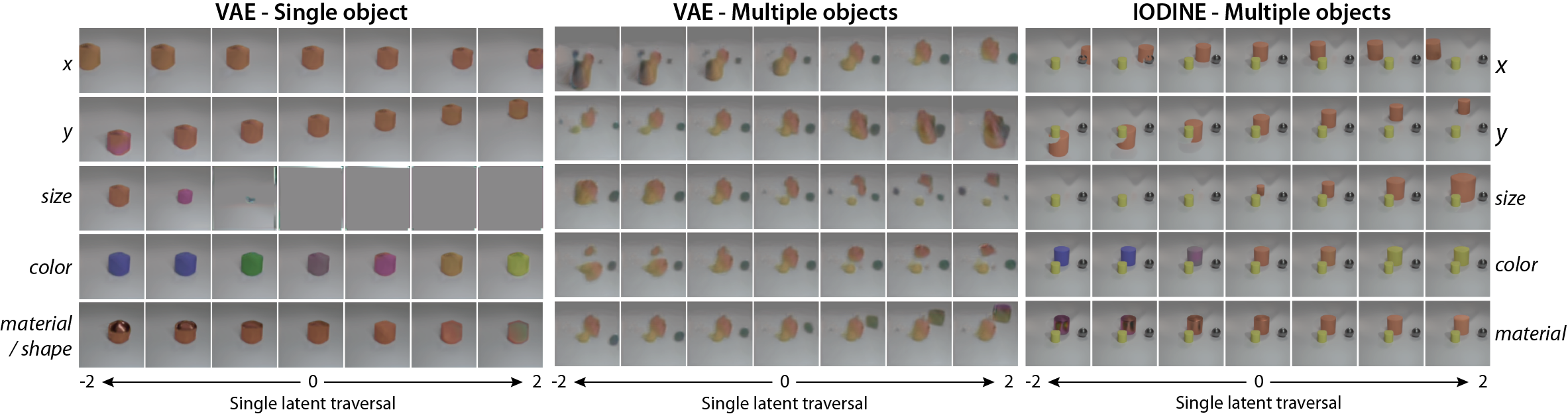}
    \vspace{-8pt}
    \caption{Disentanglement in regular VAEs vs \modelname. Rows indicate traversals of single latents, annotated by our interpretation of their effects.
    (Left) When a VAE is trained on single-object scenes it can disentangle meaningful factors of variation. 
    (Center) When the same VAE is trained on multi-object scenes, the latents entangle across both factors and objects. 
    (Right) In contrast, traversals of individual latents in \modelname vary individual factors of single objects, here the orange cylinder. Thus, the architectural bias for discovering multiple entities in a common format enables not only the discovery of objects, but also facilitates disentangling of their features.}
    \label{fig:traversals}
    \vspace{-5pt}
\end{figure*}

We evaluate our model on three main datasets: 1) CLEVR \cite{johnson2017clevr} and a variant CLEVR6 which uses only scenes with up to 6 objects, 2) a multi-object version of the dSprites dataset \cite{matthey2017dsprites}, and 3) a dataset of multiple ``Tetris''-like pieces that we created.
In all cases we train the system using the Adam optimizer \cite{kingma2015adam} to minimize the negative ELBO for $10^6$ updates.
We varied several hyperparameters, including: number of slots, dimensionality of $\z_k$, number of inference iterations, number of convolutional layers and their filter sizes, batch size, and learning rate. 
For details of the models and hyperparameters refer to \cref{appendix:model} and the code for all experiments can be found on \href{https://github.com/deepmind/deepmind-research/tree/master/iodine}{GitHub}.

\subsection{Decomposition}
\label{sec:results:segmentation}
\modelname can provide a readily interpretable segmentation of the data, as seen in \cref{fig:decompositions}.
These examples clearly demonstrate the models ability to segmenting out the same objects which were used to generate the dataset, despite never having received supervision to do so. 
To quantify segmentation quality, we measure the similarity between ground-truth (instance) segmentations and our predicted object masks using the Adjusted Rand Index (ARI; \citealt{rand1971objective, hubert1985comparing}). 
ARI is a measure of clustering similarity that ranges from $0$ (chance) to $1$ (perfect clustering) and can handle arbitrary permutations of the clusters.
We apply it as a measure of instance segmentation quality by treating each foreground pixel (ignoring the background) as one point and its segmentation as cluster assignment. 
As shown in \cref{table:baselines}, \modelname achieves almost perfect ARI scores of around $0.99$ for CLEVR6, and Tetris as well as a relatively good score of $0.77$ for Multi-dSprites.
The lower scores on Multi-dSprites are largely because \modelname struggles to produce sharp boundaries for the sprites, and we are uncertain as to the reasons for this behaviour.

We compare with MONet \cite{burgess2019monet}, following the CLEVR model implementation described in the paper except using fewer (7) slots and different standard deviations for the decoder distribution (0.06 and 0.1 for $\sigma_{\text{bg}}$ and $\sigma_{\text{fg}}$, respectively), which gave better scores.
With this, MONet obtained a similar ARI score ($0.96$) as \modelname on CLEVR6, and on Multi-dSprites it performed significantly better with a score of $0.90$ (using the unmodified model).
We also attempted to compare ARI scores to Neural Expectation Maximization, but neither Relational-NEM nor the simpler RNN-NEM variant could cope well with colored images. 
As a result, we could only compare with those methods on a binarized version of Multi-dSprites and the Shapes dataset. 
These scores are summarized in \cref{table:baselines}. 


\begin{table}
\sisetup{%
    separate-uncertainty,
    table-align-uncertainty=true,
    table-format=1.3(3),
}
\scriptsize
\centering
    \begin{tabular}{l S S S}
    \toprule
     & {IODINE} & {R-NEM} & {MONet}\\
    \midrule
    CLEVR6 & 0.988 (0) & $*$ &  0.962 (6)\\
    M-dSprites & 0.767 (56) & $*$ & 0.904 (08)  \\ 
    M-dSprites bin. & 0.648 (172) & 0.685 (17) & \\ 
    Shapes & 0.910 (119) & 0.776 (19) & \\ 
    Tetris & 0.992 (4) & $*$ & \\
    \bottomrule
    \end{tabular}
    \caption{Summary of \modelname's segmentation performance in terms of ARI (mean $\pm$ stddev across five seeds) versus baseline models. For each independent run, we computed the ARI score over 320 images, using only foreground pixels. We then picked the best hyperparameter combination for each model according to the mean ARI score over five random seeds. 
    }\label{table:baselines}
    \vspace{-5pt}
\end{table}

\subsection{Representation Quality}
\label{sec:results:representation}
\paragraph{Information Content}

The object-reconstructions in \cref{fig:decompositions} show that their representations contain all the information about the object.
But in what format, and how usable is it? 
To answer this question we associate each ground-truth object with its corresponding $\z_k$ based on the segmentation masks.
We then train a single-layer network to predict ground-truth factors for each object.
Note that this predictor is trained \emph{after} \modelname has finished training (i.e. no supervised fine-tuning).
It tells us if a linear mapping is sufficient to extract information like color, position, shape or size of an object from its latent representation, and gives an important indication about the usefulness of the representation.
Results in \cref{fig:factors} clearly show that a linear mapping is sufficient to extract relevant information about these object attributes from the latent representation to high accuracy.
This result is in contrast with the scene representations learned by a standard VAE.
Here even training the factor-predictor is difficult, as there is no obvious way to align objects with features.
To make this comparison, we chose a canonical ordering of the objects based on their size, material, shape, and position (with decreasing precedence). The precedence of features was intended as a heuristic to maximize the predictability of the ordering. We then trained a linear network to predict the concatenated features of the canonically ordered objects from the latent scene representation.
As the results in \cref{fig:factors} indicate, the information is present, but in a much less explicit/usable state.

\paragraph{Disentanglement}

Disentanglement is another important desirable property of representations \cite{bengio2013representation} that captures how well learned features separate and correspond to individual, interpretable factors of variation in the data.
While its precise definition is still highly debated \cite{higgins2018definedis, eastwood2018quantDisent, ridgeway2018fstatisticDisent,locatello2018challenging}, the concept of disentanglement has generated a lot of interest recently.
Good disentanglement is believed to lead to both better generalization and more interpretable features \cite{lake2016learnLikePeople, higgins2017darla}.
Interestingly, for these desirable advantages to bear out, disentangled features seem to be most useful for properties of single objects, such as color, position, shape, etc. It is much less clear how to operationalize this in order to create disentangled representations of entire scenes with variable numbers of objects.
And indeed, if we train a VAE that can successfully disentangle features of a single-object dataset, we find that that its representation becomes highly entangled on a multi-object dataset, (see \cref{fig:traversals} left vs middle). 
\modelname, on the other hand, successfully learns disentangled representations, because it is able to first decompose the scene and then represent individual objects
(\cref{fig:traversals} right).
In \cref{fig:traversals} we show traversals of the most important features (selected by KL) of a standard VAE vs \modelname. 
While the standard VAE clearly entangles many properties even across multiple objects, \modelname is able to neatly separate them.


\paragraph{Generalization}
\label{sec:results:generalization}
\begin{figure}[t]
    \centering
    \includegraphics[width=0.98\columnwidth]{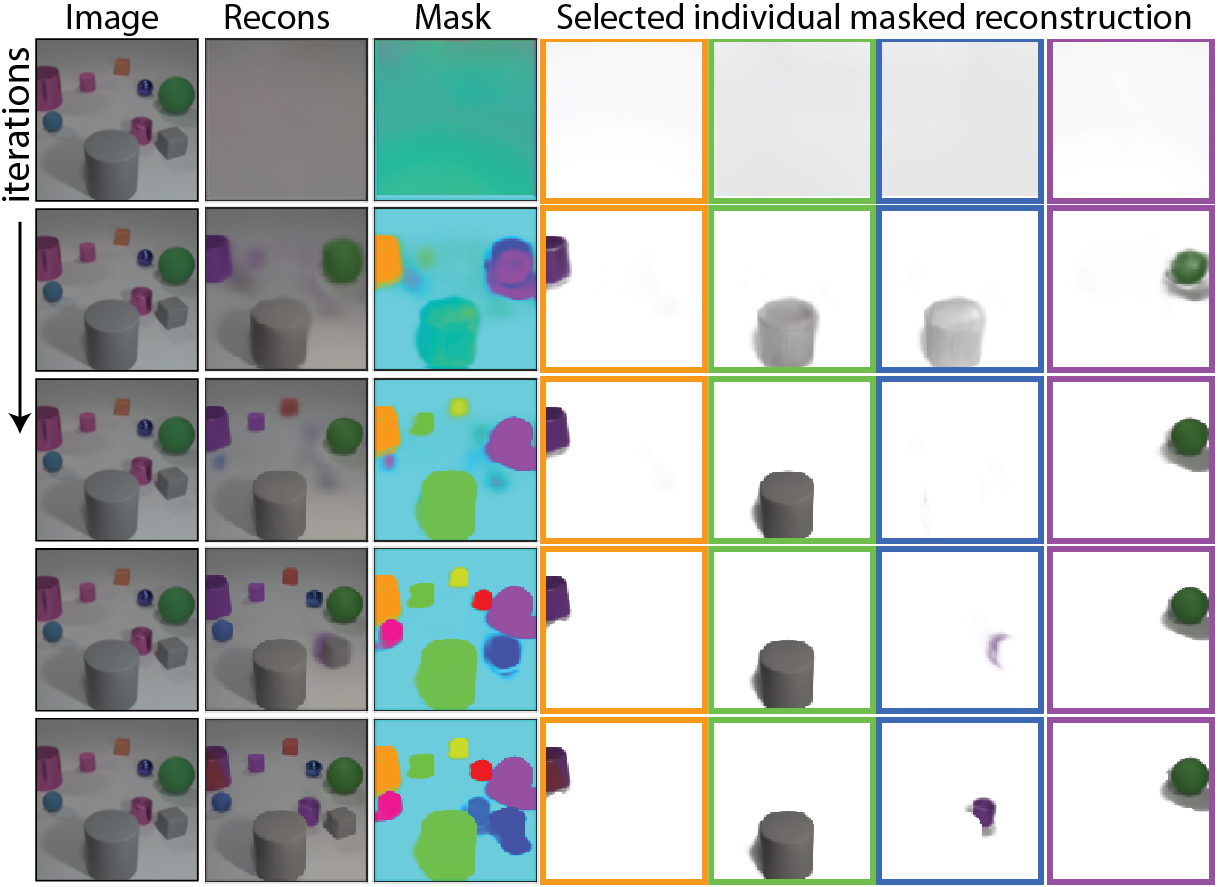}
    \vspace{-5pt}
    \caption{\modelname's iterative inference process and generalization capabilities.
    Rows indicate steps of iterative inference, refining reconstructions and segmentations when moving down the figure.
    Of particular interest is the \emph{explaining away} effect visible between slots 2 and 3, where they settle on different objects despite both starting with the large cylinder.
    The model was only trained with $K=7$ slots on 3-6 objects (excluding green spheres), and yet is able to generalize to $K=11$ slots (only 4 are shown, see \cref{fig:generalize_color_full} in the appendix for a full version) on a scene with $9$ objects, including the never seen before green sphere (last column).
    }
    \label{fig:generalize_objs}
    \label{fig:generalize_color}
\end{figure}
Finally, we can ask directly: Does the system generalize to novel scenes in a systematic way?
Specifically, does it generalize to scenes with more or fewer objects than ever encountered during training? 
Slots are exchangeable by design, so we can freely vary the number of slots during test-time (more on this in \cref{sec:results:robustness}).
So in \cref{fig:generalize_objs} we qualitatively show the performance of a system that was trained with $K=7$ on up to 6 objects, but evaluated with $K=11$ on 9 objects.
In \cref{fig:slots:ari} the orange boxes show, that, even quantitatively, the segmentation performance decreases little when generalizing to more objects.

A more extreme form of generalization involves handling unseen feature combinations. 
To test this we trained our system on a subset of CLEVR that does not contain green spheres (though it does contain spheres and other green objects).
And then we tested what the system does when confronted with a green sphere.
In \cref{fig:generalize_color} it can be seen that \modelname is still able to represent green spheres, despite never having seen this combination during training.

\subsection{Robustness \& Ablation}
\label{sec:results:robustness}

\paragraph{Iterations}
\begin{figure}
    \begin{subfigure}[b]{0.32\columnwidth}
        \includegraphics[width=\columnwidth]{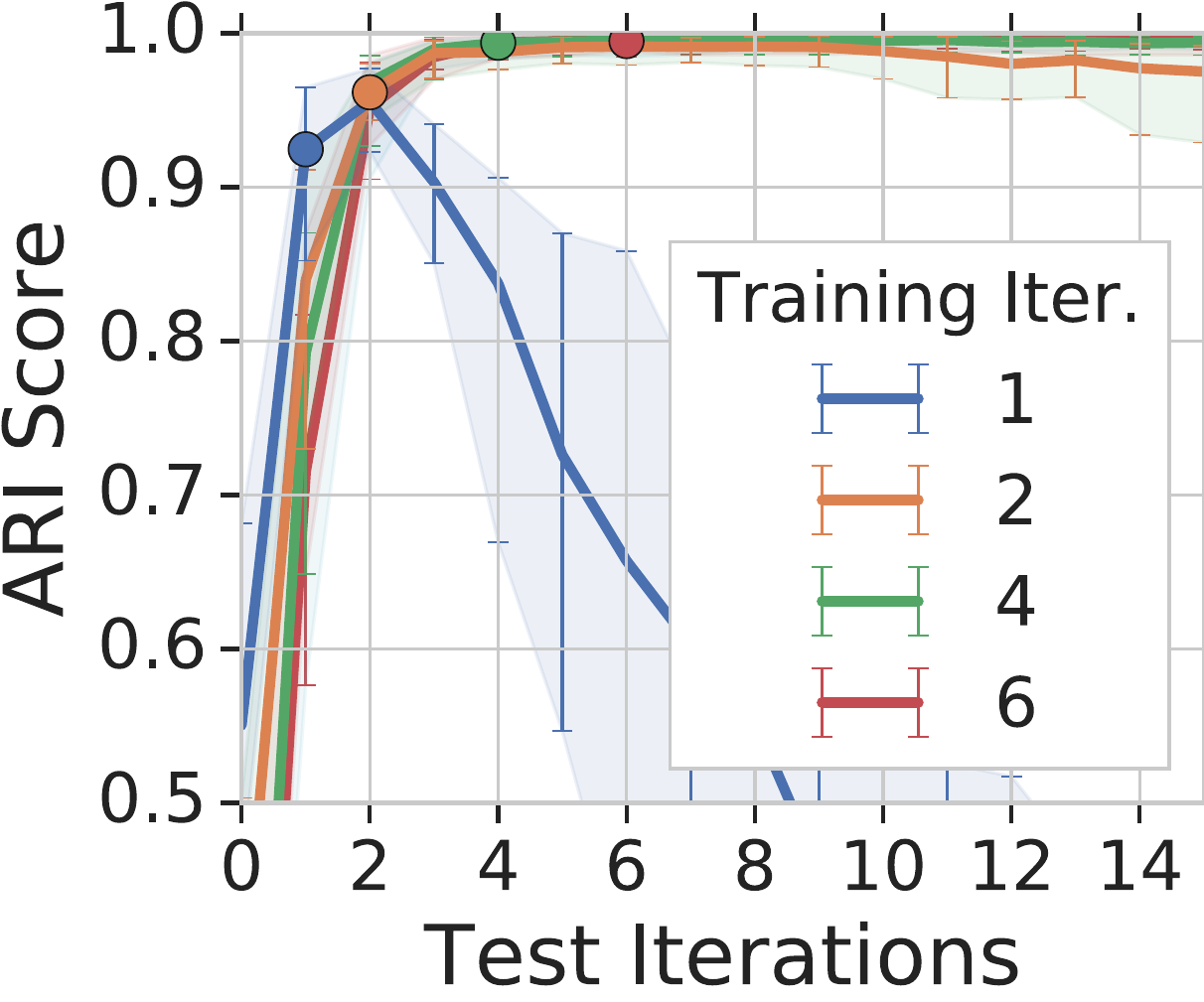}
        \caption{ARI}
        \label{fig:iterations:ari}
    \end{subfigure}
    \begin{subfigure}[b]{0.32\columnwidth}
        \includegraphics[width=\columnwidth]{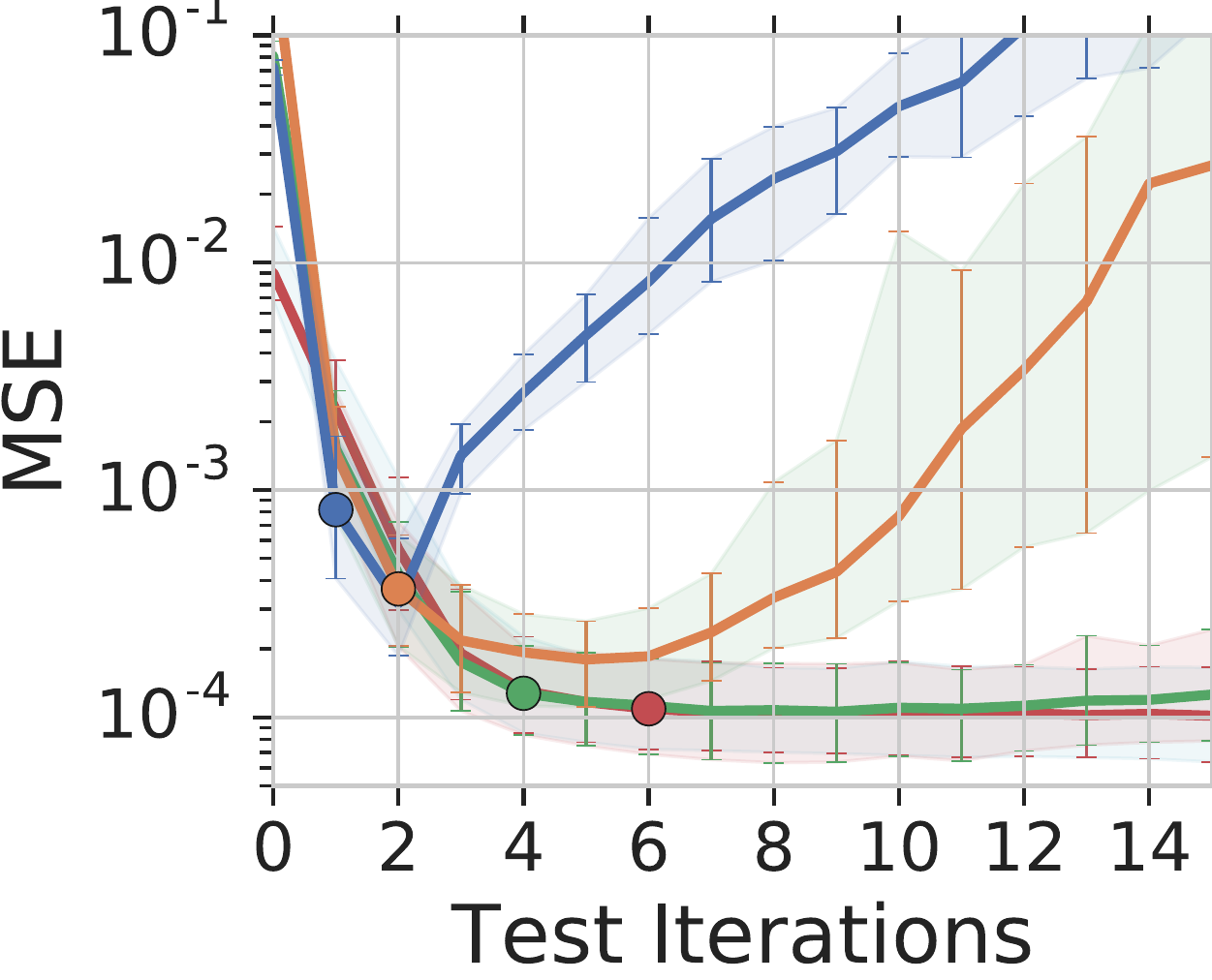}
        \caption{MSE}
        \label{fig:iterations:mse}
    \end{subfigure}
    \begin{subfigure}[b]{0.32\columnwidth}
        \includegraphics[width=\columnwidth]{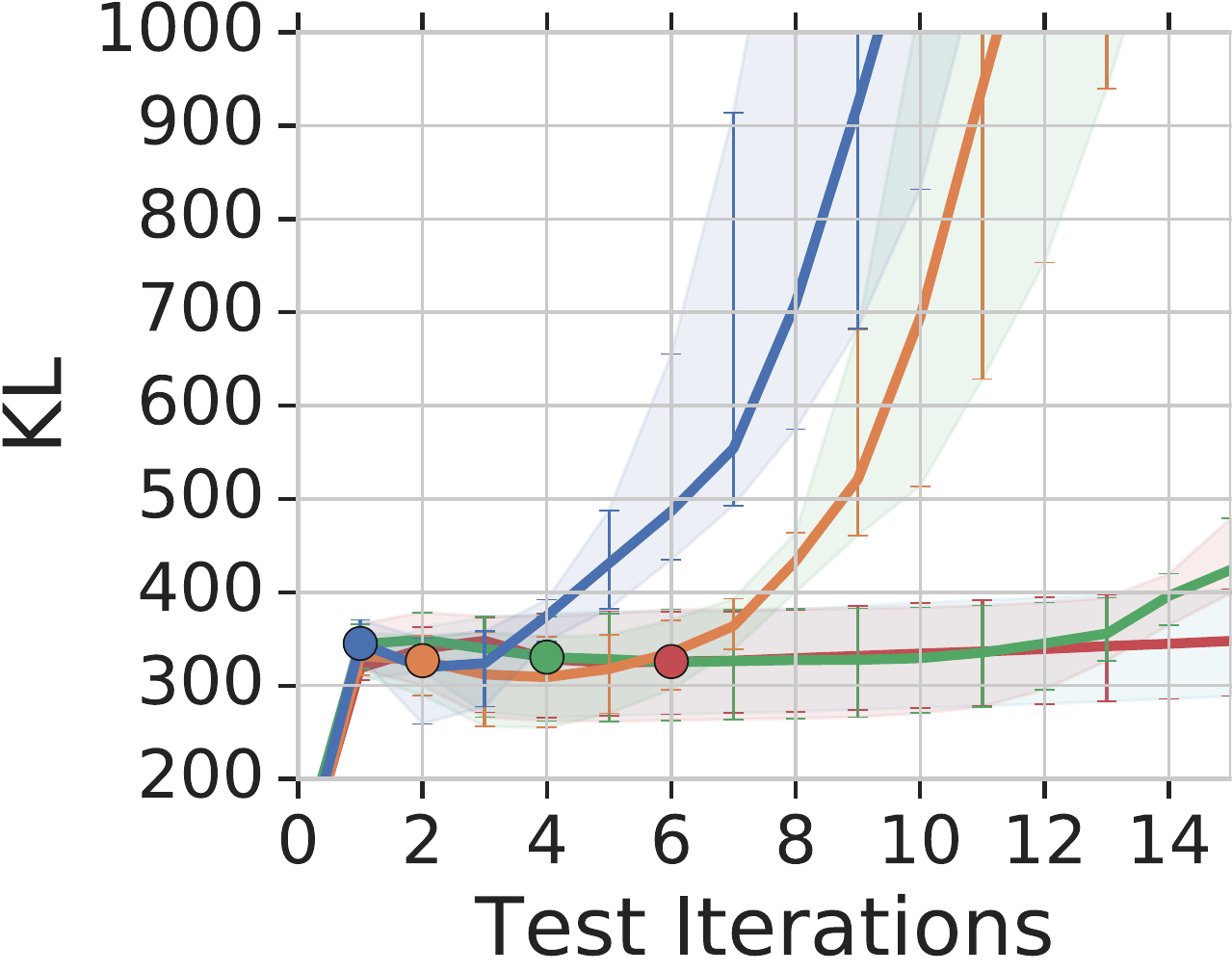}
        \caption{KL}
        \label{fig:iterations:kl}
    \end{subfigure}
    \caption{The effect of varying the number of iterations, for both training and at test time. (a) Median ARI score, (b) MSE and (c) KL over test-iterations, for models trained with different numbers of iterations on CLEVR6.
    The region beyond the filled dots thus shows test-time generalization behavior.
    Shaded region from 25th to 75th percentile.}
    \label{fig:iterations}
\end{figure}

The number of iterations is one of the central hyperparameters to our approach.
To investigate its impact, we trained four models with 1, 2, 4 and 6 iterations on CLEVR6, and evaluated them all using 15 iterations (c.f. \cref{fig:iterations}).
The first thing to note is that the inference converges very quickly within the first 3-5 iterations after which neither the segmentation nor reconstruction change much.
The second important finding is that the system is very stable for much longer than the number of iterations it was trained with. 
The model even further improves the segmentation and reconstruction when it is run for more iterations, though it eventually starts to diverge after about two to three times the number of training iterations as can be seen with the blue and orange curves in \cref{fig:iterations}.

\paragraph{Slots}

\begin{figure}[t]
     \begin{subfigure}[b]{0.32\columnwidth}
        \includegraphics[width=\columnwidth]{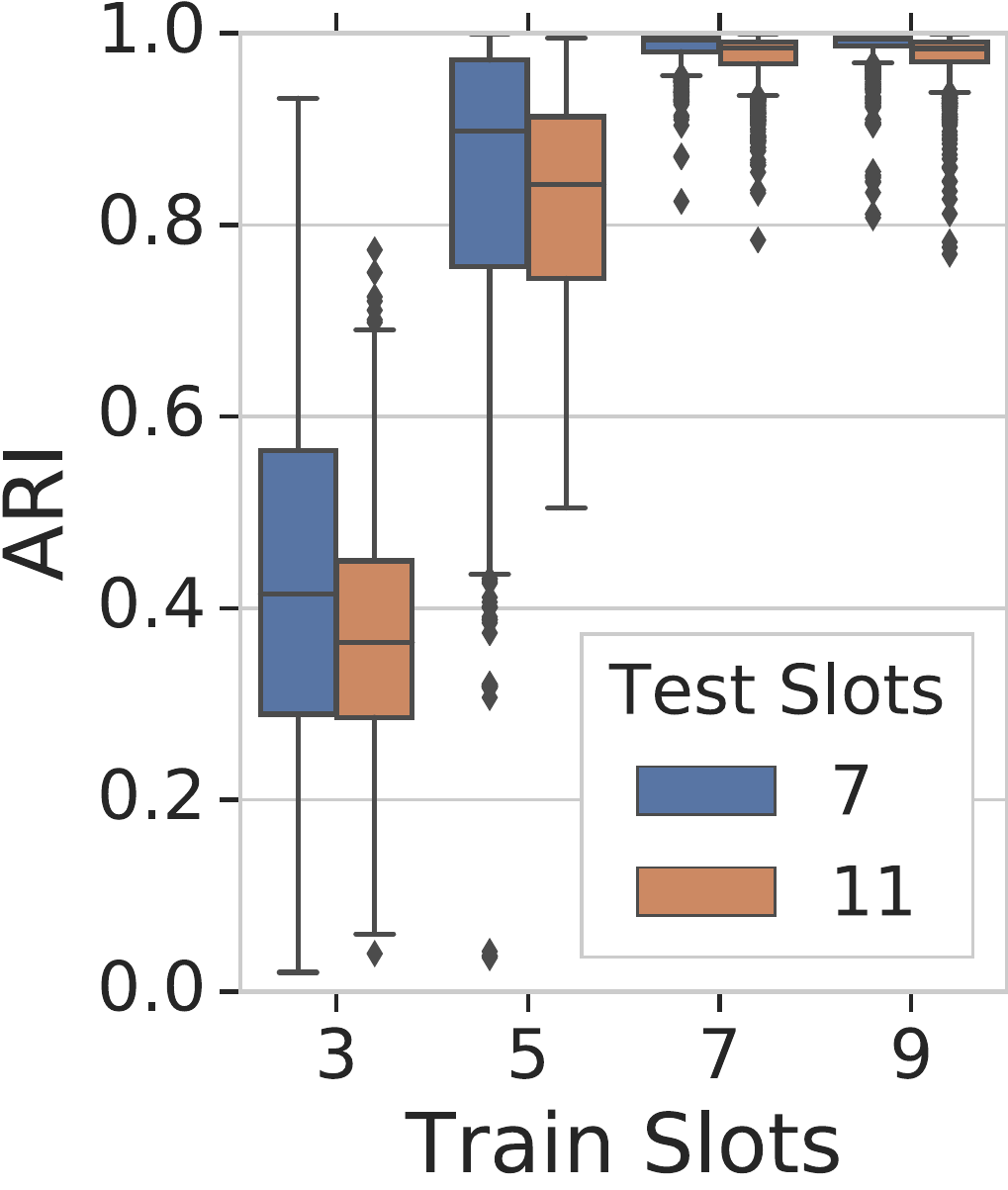}
        \caption{ARI}
        \label{fig:slots:ari}
    \end{subfigure}
    \begin{subfigure}[b]{0.32\columnwidth}
        \includegraphics[width=\columnwidth]{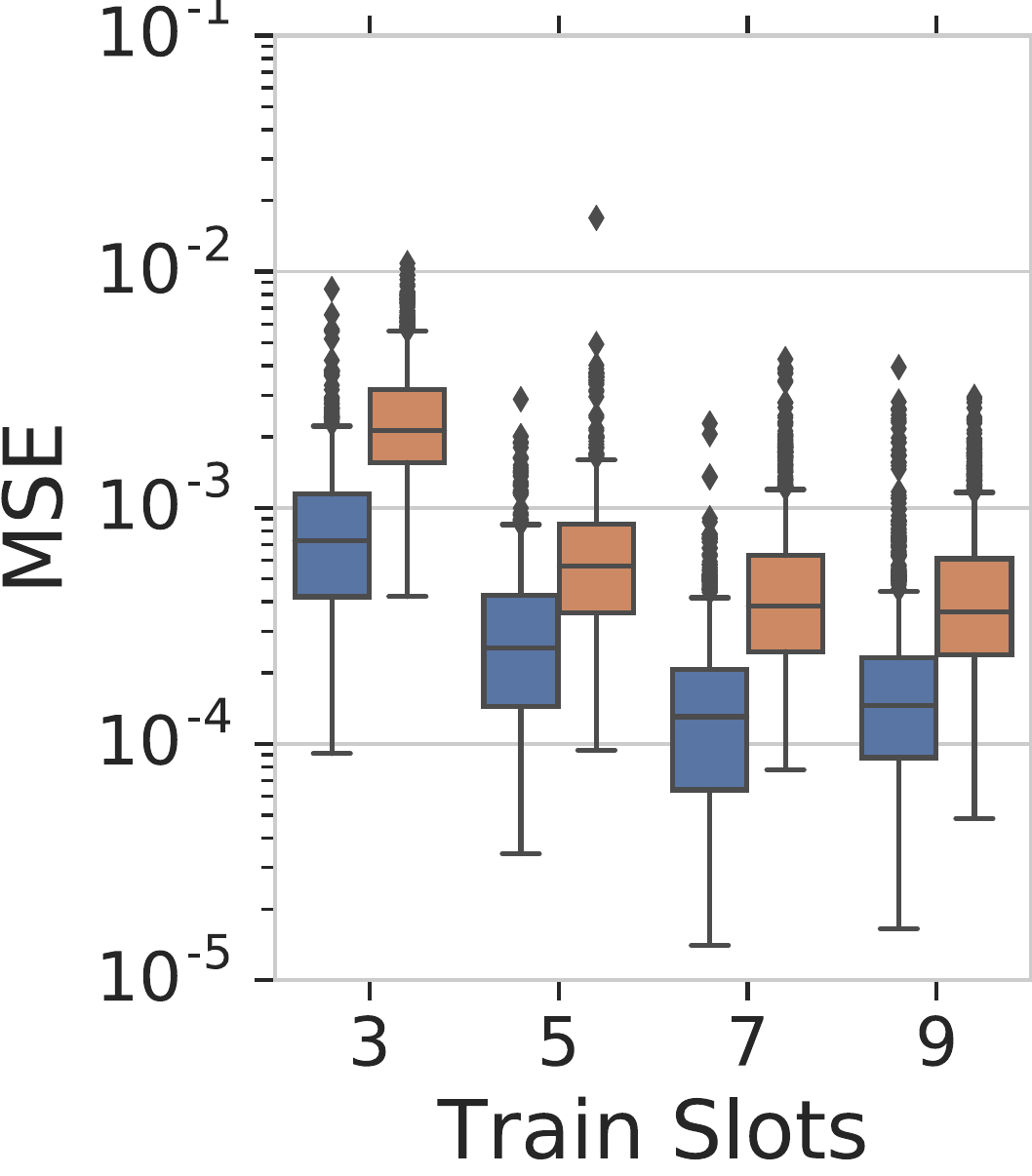}
        \caption{MSE}
        \label{fig:slots:mse}
    \end{subfigure}
    \begin{subfigure}[b]{0.32\columnwidth}
        \includegraphics[width=\columnwidth]{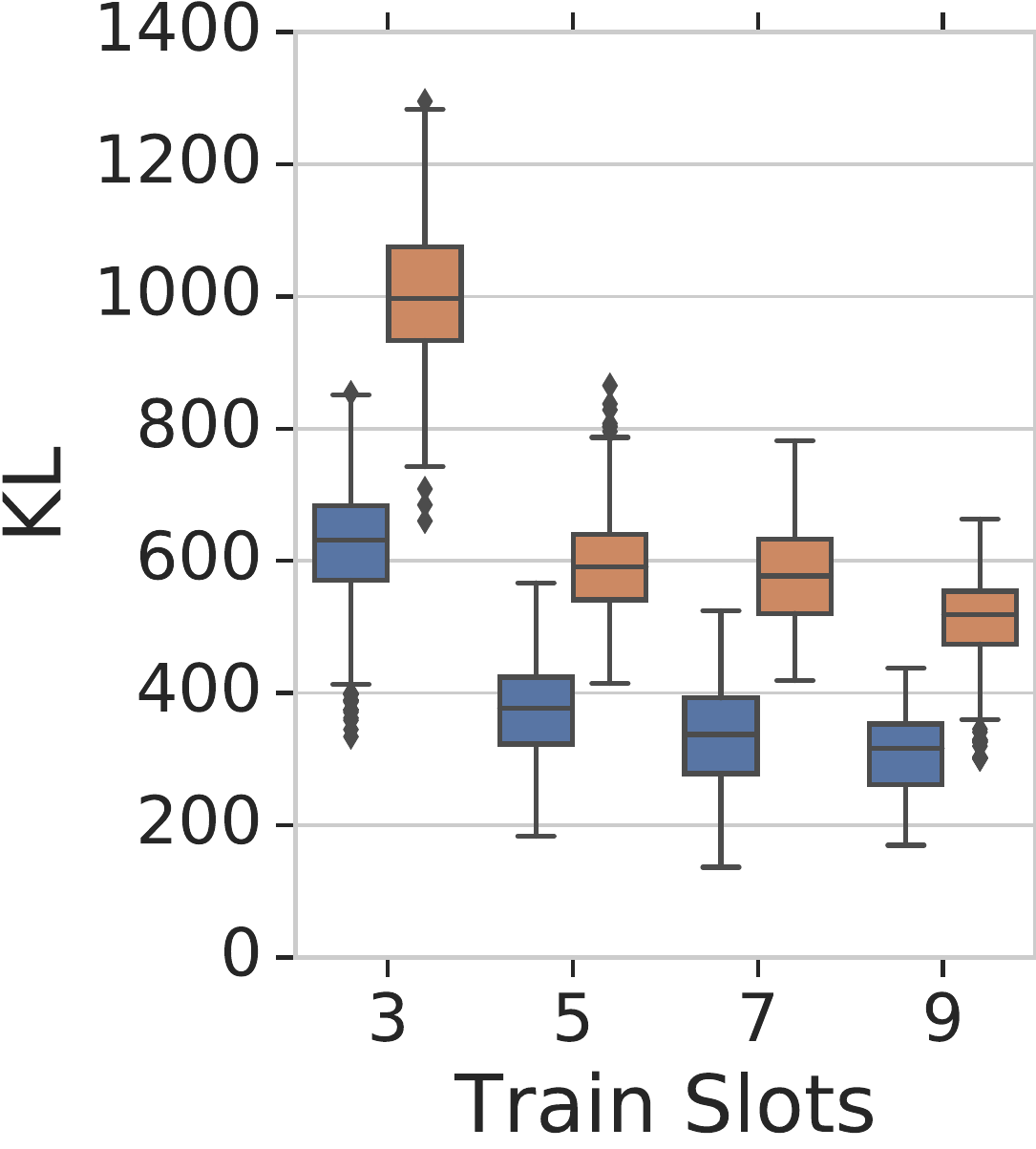}
        \caption{KL}
        \label{fig:slots:kl}
    \end{subfigure}
    \vspace*{-5pt}
    \caption{\modelname trained on CLEVR6 with varying numbers of slots (columns). Evaluation of (a) ARI, (b) MSE, and (c) KL with 7 slots on 3-6 Objects (blue) and 11 slots on 3-9 objects (orange).}
    \label{fig:slots}
    \vspace*{-10pt}
\end{figure}

The other central parameter of \modelname is the number of slots $K$, as it controls the maximum number of objects the system can separate.
It is important to distinguish varying $K$ for training vs varying it at test-time.
As can be seen in \cref{fig:slots}, if the model was trained with sufficiently many slots to fit all objects ($K=7$, and $K=9$), then test-time behavior generalizes very well. 
Typical behavior (not shown) is to leave excess slots empty, and when confronted with too many objects it will often completely ignore some of them, leaving the other object-representations mostly intact.
Given enough slots at test time, such a model can even segment and represent scenes of higher complexity (more objects) than any scene encountered during training (see \cref{fig:generalize_objs} and the orange boxes in \cref{fig:slots}).
If on the other hand, the model was trained with too few slots ($K=3$ and $K=5$), its performance suffers substantially.
This happens because, here the only way to reconstruct the entire scene during training is to consistently represent multiple objects per slot. 
And that leads to the model learning inefficient and entangled representations akin to the VAE in \cref{fig:traversals} (also apparent from their much higher KL in \cref{fig:slots:kl}).
Once learned, this sub-optimal strategy cannot be mitigated by increasing the number of slots at test-time as can be seen by their decreased performance in \cref{fig:slots:ari}.

\paragraph{Input Ablations}
We ablated each of the different inputs to the refinement network described in \cref{sec:method:inputs}.
Broadly, we found that individually removing an input did not noticeably affect the results (with two exceptions noted below). See Figures \ref{fig:abl_clevr_loss}-\ref{fig:abl_tetris_kl} in the Appendix demonstrating this lack of effect on different terms of the model's loss and the ARI segmentation score on both CLEVR6 and Tetris.
A more comprehensive analysis could ablate combinations of inputs and identify synergistic or redundant groups, and thus potentially simplify the model.
We didn't pursue this direction since none of the inputs incurs any noticeable computational overhead and at some point during our experimentation each of them contributed towards stable training behavior.

The main exceptions to the above are $\nabla_\lambda\mathcal{L}$ and $\x$. 
Computing the former requires an entire backward pass through the decoder, and contributes about $20\%$ of the computational cost of the entire model. 
But we found that it often substantially improves performance and training convergence, which justifies its inclusion.
A somewhat surprising finding was that for the Tetris dataset, removing $\x$ from the list of inputs had a pronounced detrimental effect, while for CLEVR it was negligible. 

\paragraph{Broadcast Decoder Ablation}
\label{sec:results:ablation}
We use the spatial broadcast decoder~\cite{watters2019} primarily for its significant impact on the disentanglement of the representations, but its continuous spatial representation bias also seems to help decomposition.
When replacing it with a deconvolution-based decoder the factor regression scores on CLEVR6 are significantly worse as can be seen in \cref{fig:factors}.
Especially for shape and size it now performs no better than the VAE which uses spatial broadcasting.
The foreground-ARI scores also drop significantly ($0.67 \pm 0.06$ down from $0.99$) and the model seems less able to specialize slots to single objects (see \cref{fig:broadcast_ablation}).
Note though, that these discrepancies might easily be reduced, since we haven't invested much effort in tuning the architecture of the deconv-based decoder.

\subsection{Multi-Modality and Multi-Stability}
\label{sec:results:multistability}

\begin{figure}[t!]
    \centering
    \includegraphics[width=0.92\columnwidth]{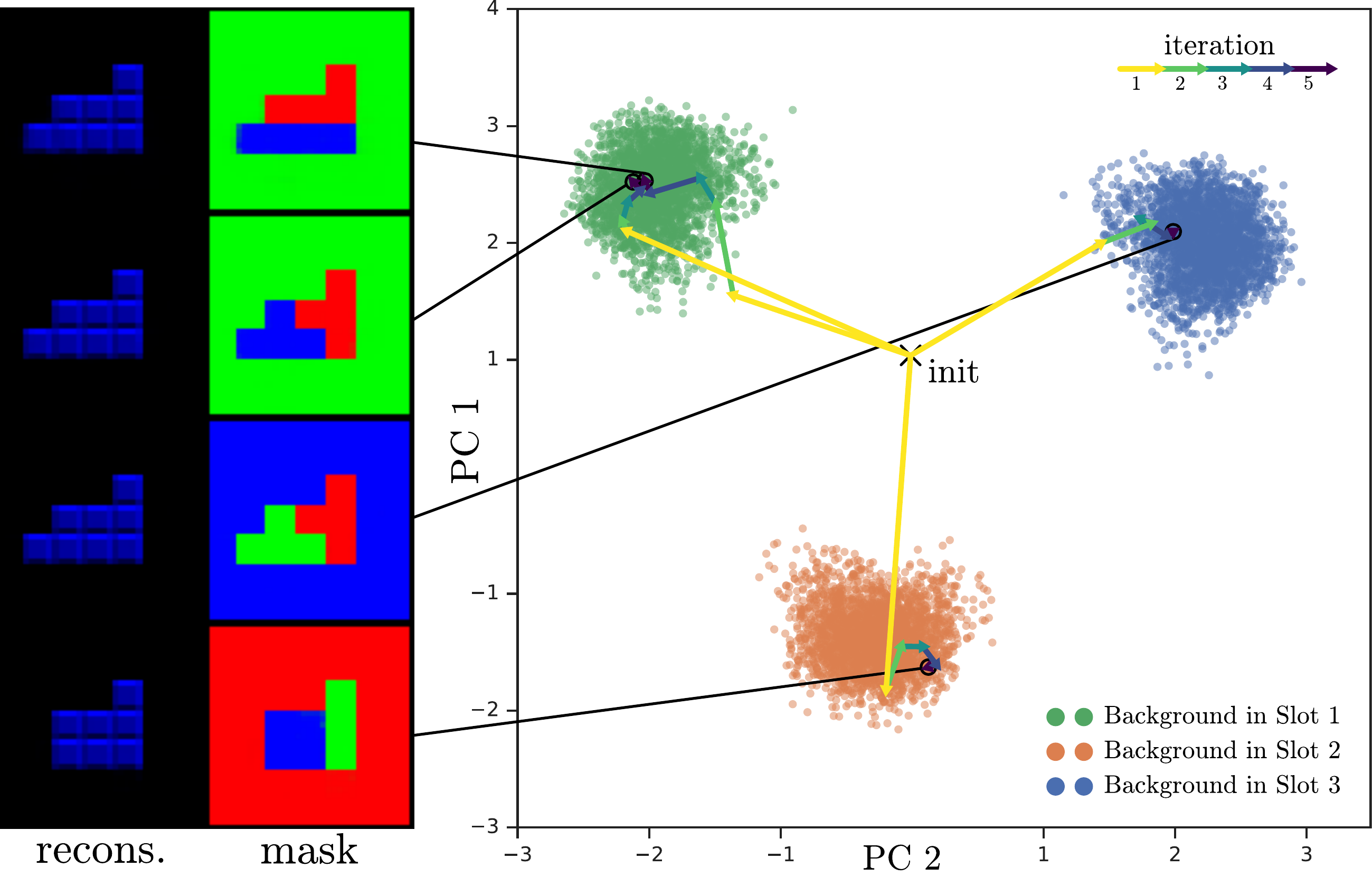}
    \vspace{-10pt}
    \caption{Multi-stability of segmentation when presented with an ambiguous stimulus. 
    \textbf{Left:} Depending on the random sampling during iterative refinement, \modelname can produce different permutations of groups (row 2 vs 3), a different decomposition (row 1) or sometimes an invalid segmentation and reconstruction (row 4).
    \textbf{Right:} PCA of the latent space, coloured by which slot corresponds to the background.
    Paths show the trajectory of the iterative refinement for the four examples on the left.}
    \label{fig:tetris}
\end{figure}

\begin{figure}[t!]
    \begin{subfigure}[b]{0.46\columnwidth}
        \includegraphics[width=\columnwidth]{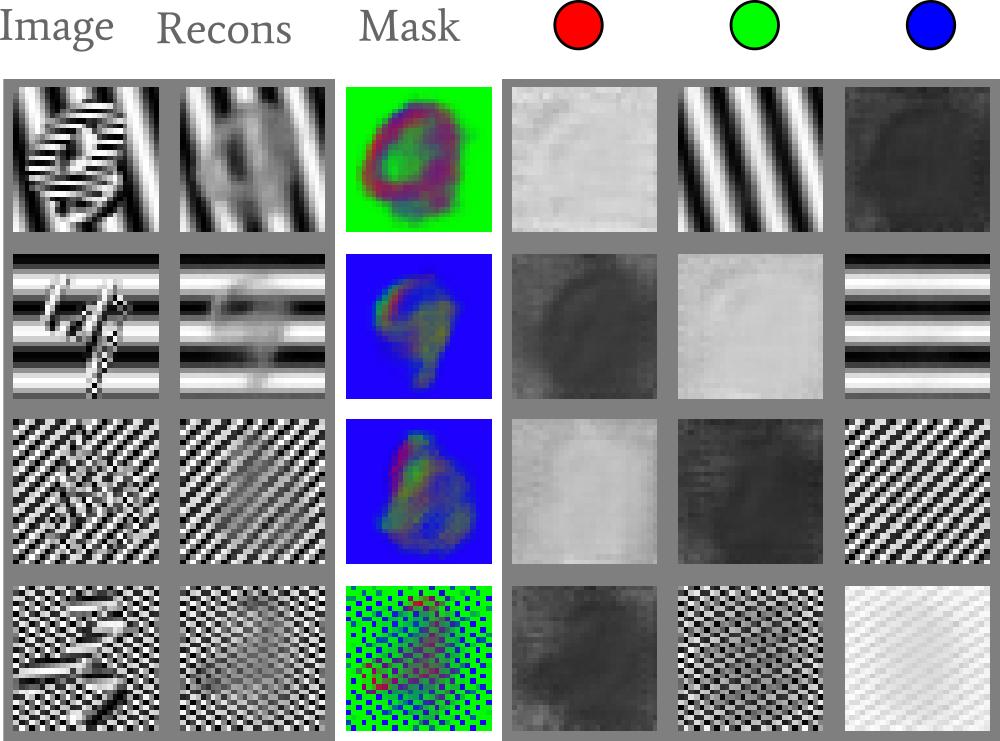}
        \caption{Textured MNIST}
        \label{fig:fail:tmnist}
    \end{subfigure}
    \begin{subfigure}[b]{0.53\columnwidth}
        \includegraphics[width=\columnwidth]{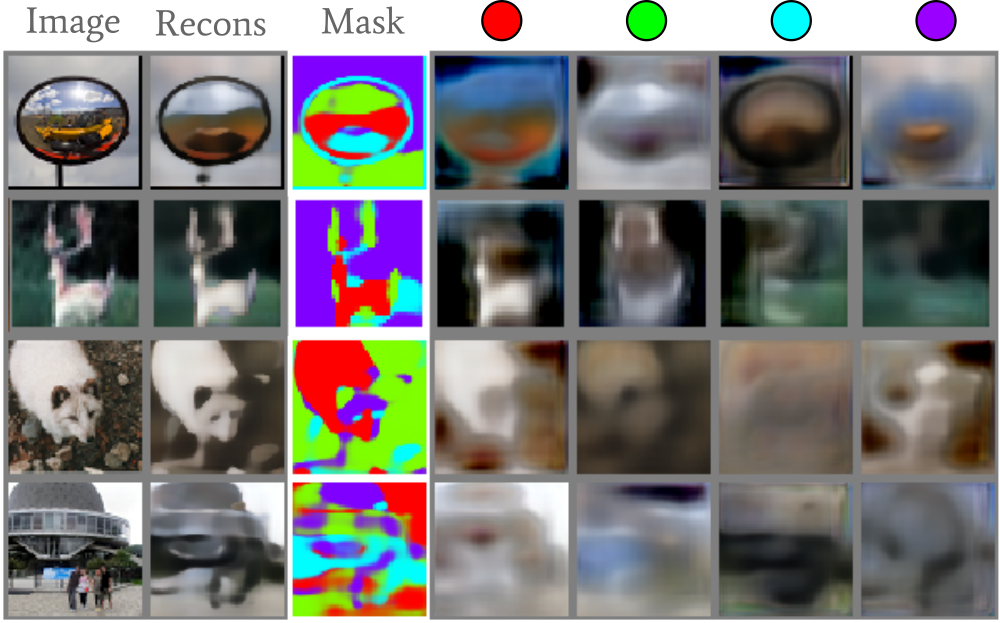}
        \caption{ImageNet}
        \label{fig:fail:imagenet}
    \end{subfigure}
    \begin{subfigure}[b]{\columnwidth}
        \includegraphics[width=\columnwidth]{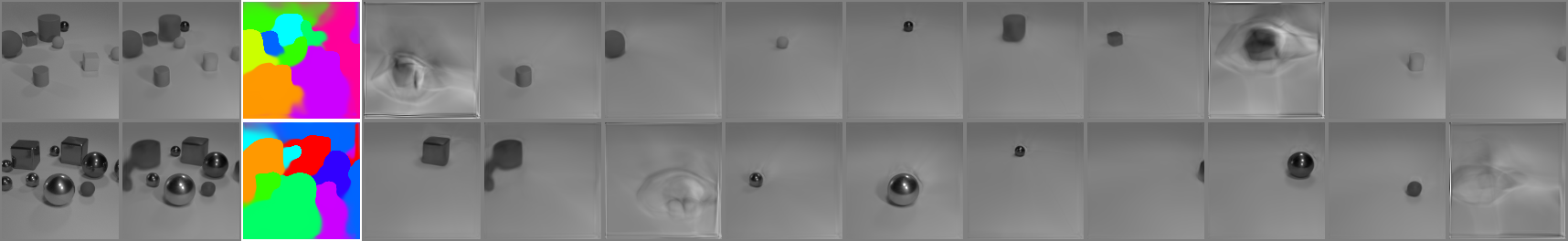}
        \caption{Grayscale CLEVR}
        \label{fig:fail:clevr}
    \end{subfigure}
    \vspace*{-20pt}
    \caption{Segmentation challenges a) \modelname did not succeed in capturing the foreground digits in the Textured MNIST dataset.
    b) \modelname groups ImageNet not into meaningful objects but mostly into regions of similar color.
    c) On a grayscale version of CLEVR, \modelname still produces the desired groupings.}\label{fig:fail}
    \vspace{-10pt}
\end{figure}

Standard VAEs are unable to represent multi-modal posteriors, because $q_\lambda(\z|\x) $ is parameterized using a unimodal Gaussian distribution.
However, as demonstrated in \cref{fig:tetris}, \modelname can actually handle this problem quite well.
How is that possible? 
It turns out that this is an important side-effect of iterative variational inference, that to the best of our knowledge has not been noticed before:
The stochasticity at each iteration, which results from sampling \z to approximate the likelihood, implicitly acts as an auxilliary (inference) random variable.
This effect compounds over iterations, and is amplified by the slot-structure and the effective message-passing between slots over the course of iterations.
In effect the model can implicitly represent multiple modes (if integrated over all ways of sampling \z) and thus converge to different modes (see \cref{fig:tetris} left) depending on these samples.
This does not happen in a regular VAE, where no stochasticity enters the inference process.
If we had an exact and deterministic way to compute the likelihood and its gradient, this effect would vanish.

A neat side-effect of this is the ability to elegantly capture ambiguous (aka multi-stable) segmentations such as the ones shown in \cref{fig:tetris}.
We presented \modelname with an ambiguous arrangement of Tetris blocks, which has three different yet equally valid "explanations" (given the data distribution).
When we evaluate a trained model on this image, we get different segmentations on different evaluations.
Some of these correspond to different slot-orderings (1st vs 3rd row).
But we also find qualitatively different segmentations (i.e. 3rd vs 4th row) that correspond to different interpretations of the scene.
This is an impressive result given that multi-stability is a well-studied, pervasive feature of human perception that is important for handling ambiguity, and that is not modelled by any standard image recognition networks.

\section{Discussion and Future Work}
\label{sec:discussion}

We have introduced IODINE, a novel approach for unsupervised representation learning of multi-object scenes, based on amortized iterative refinement of the inferred latent representation.
We analyzed IODINE’s performance on various datasets, including realistic images containing variable numbers of partially occluded 3D objects, and demonstrated that our method can successfully decompose the scenes into objects and represent each of them in terms of their individual properties such as color, size, and material. 
IODINE can robustly deal with occlusions by inpainting covered sections, and generalises beyond the training distribution in terms of numerosity and object-property combinations.
Furthermore, when applied to scenes with ambiguity in terms of their object decomposition, \modelname can represent -- and converge to -- multiple valid solutions given the same input image.


We also probed the limits of our current setup by applying \modelname to the Textured MNIST dataset~\cite{greff2016tagger} and to ImageNet, testing how it would deal with texture-segmentation and more complex real-world data (\cref{fig:fail}).
Trained on ImageNet data, \modelname segmented mostly by color rather than by objects. 
This behavior is not unexpected: ImageNet was never designed as a dataset for unsupervised learning, and likely lacks the richness in poses, lighting, sizes, positions and distance variations required to learn object segmentations from scratch. 
Trained on Textured MNIST, \modelname was able to model the background, but mostly failed to capture the foreground digits.
Together these results point to the importance of color as a strong cue for segmentation, especially early in the iterative refinement process.
As demonstrated by our results on grayscale CLEVR (\cref{fig:fail:clevr}) though, color is not a requirement. 

Beyond more diverse training data, we want to highlight three other promising directions to scale \modelname to richer real-world data.
First, an extension to sequential data is attractive, because temporal data naturally contains rich statistics about objectness both in the movement itself, and in the smooth variations of object factors.
\modelname can readily be applied to sequences feeding a new frame at every iteration, and we have done some preliminary experiments described in \cref{appendix:discussion:sequences}.
As a nice side-effect, the model automatically maintains the object to slot association, turning it into an unsupervised object tracker.

Physical interaction between objects is another common occurrence in sequential data. \modelname in its current form has limited abilities for modelling dynamics. Even statically placed objects commonly adhere to certain relations between each other, such as cars on streets.
\modelname currently assumes objects to be placed independently of each other; relaxing this assumption will be important for modelling physical interactions. 
Yet there is also a need to balance this with the independence assumption required to split objects, since the system should still be able to segment out a car floating in space. 
Thus we believe integration with some form of graph network to support relations while preserving slot symmetry is another promising direction.

Finally, object representations have to be useful, such as for supervised tasks, or for agents in reinforcement learning setups.
Whatever the task, it should provide important feedback about which objects matter and which are irrelevant.
Complex visual scenes can contain an extremely large number of potential objects (think of sand grains on a beach), which can make it unfeasible to represent them all simultaneously.
Allowing task-related signals to bias selection, for what and how to decompose, may enable scaling up unsupervised scene representation learning approaches like \modelname to arbitrarily complex scenes. 

\section*{Acknowledgements}
We would like to thank Danilo Rezende, Sjoerd van Steenkiste, and Malcolm Reynolds for helpful suggestions and generous support.

\bibliography{main}
\bibliographystyle{icml2019}

\appendix
\section{Further Discussion}
\label{appendix:discussion}

\subsection{Sequences}
\label{appendix:discussion:sequences}

\begin{figure}[h!]
    \centering
    \includegraphics[width=0.98\columnwidth]{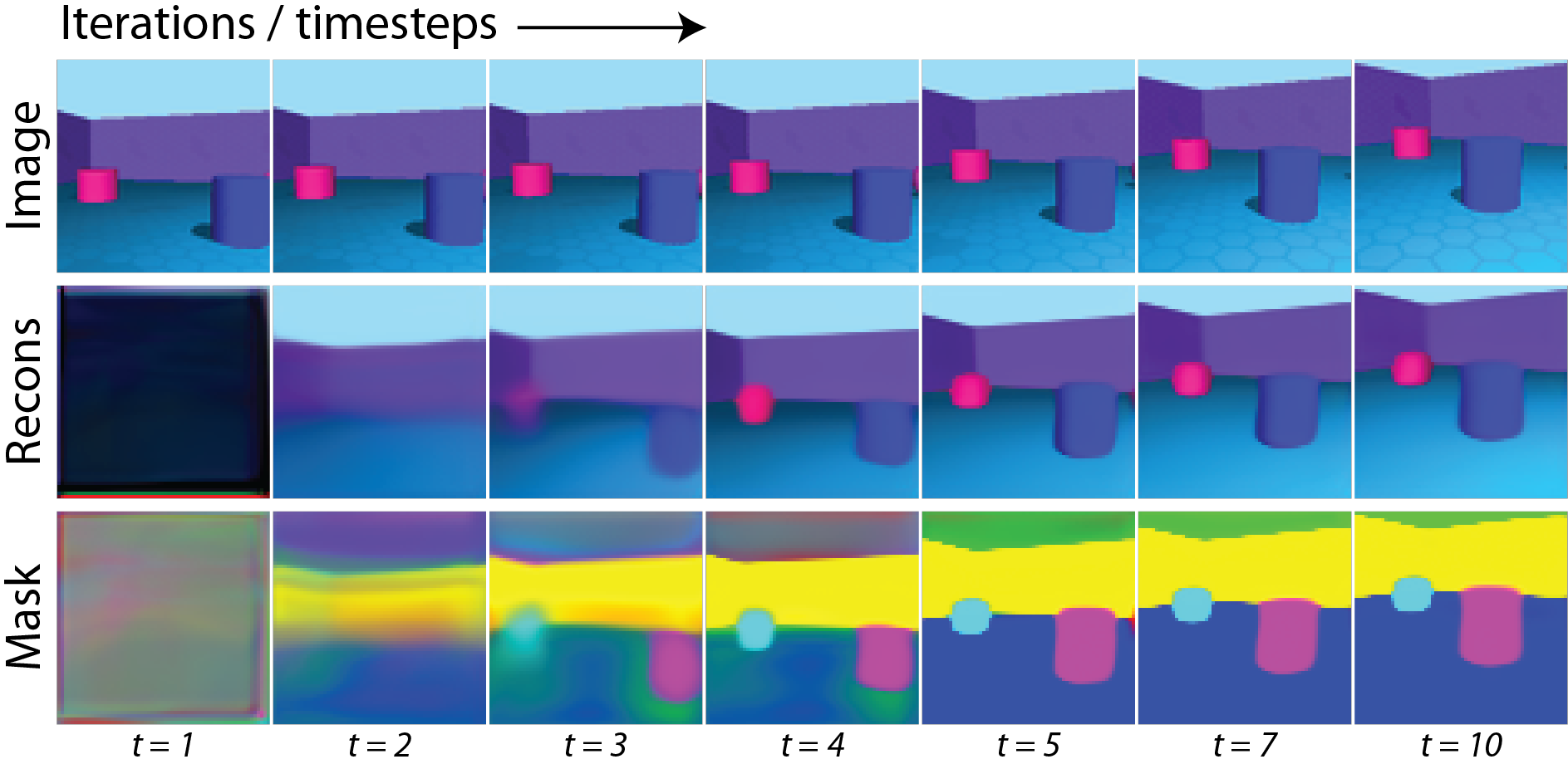}
    \vspace*{-10pt}
    \caption{\modelname applied to Objects Room sequences by setting $N$, the number of refinement iterations, equal to the number of timesteps in the data.}
    \label{fig:sequence}
\end{figure}

The iterative nature of \modelname lends itself readily to sequential data, by, e.g., feeding a new frame at every iteration, instead of the same input image $\x$.
This setup corresponds to one iteration per timestep, and using next-step-prediction instead of reconstruction as part of the training objective.
An example of this can be seen in \cref{fig:sequence} where we show a 16 timestep sequence along with reconstructions and masks.
When using the model in this way, it automatically maintains the association of object to slot over time (i.e, displaying robust slot stability). 
Thus, object tracking comes almost for free as a by-product in \modelname. 
Notice though, that \modelname has to rely on the LSTM that is part of the inference network to model any dynamics.
That means none of the dynamics of tracked objects (e.g. velocity) will be part of the object representation.

\subsection{Memory Limitations}
\label{appendix:discussion:memory}
It is worth pointing out that memory consumption presents an important limiting factor to scaling \modelname.
To allow training by backpropagation, each slot and each refinement step require the storage of activations for an entire decoder and refinement network.
Memory consumption during training thus scales linearly with both $K$ and $T$.
This is particularly restrictive for sequential data, where the number of steps can grow very large.
In our experiments from \cref{appendix:discussion:sequences}, we found that 16 timesteps with a batch-size of 4 was the upper limit on V100 GPUs with 16GB of RAM.
Of course this also depends on the size of the input and the size of the network. 
Note also that at inference time there is no need to keep the activations of previous timesteps, so the dependence on $T$ can be eliminated there.

\subsection{Comparison with MONet}
\label{appendix:discussion:monet}
The Multi-Object NETwork (MONet; \citealt{burgess2019monet}) is a complementary method for unsupervised object representation learning also developed recently.
It learns to sequentially attend to individual objects using a masking network and a VAE. 
In each step the masking network segments out a yet unexplained part of the image (the next object) which is then fed to the VAE which has to reconstruct that object and the mask.
Thus, in contrast to \modelname, MONet uses one iteration per object and doesn't adjust an object once it has been covered.

Both methods focus on the representation learning aspect and both ensure that all objects are encoded in the same format by sharing weights across objects.
In our preliminary experiments MONet produced results very similar to \modelname on CLEVR both in terms of segmentation and regarding the quality of object representations, and also learns to inpaint occluded parts of objects. 

Since MONet only visits each object once, it is a more lightweight method that requires less computation and memory to train and run.
Recurrently iterating over objects also has the benefit that the model can dynamically vary the number of objects, whereas in \modelname the maximum number of objects is a hyperparameter that has to be fixed manually (though it can be changed at test time). 
The usage of a separate masking network which isn't directly subject to a representational bottleneck likely leads to less regularization for the segmentation mask.
This could potentially allow MONet to better deal with complex segmentation shapes. 
But it also has to use that ability to directly produce masks that respect occlusion, whereas \modelname tends to produce masks for full unoccluded objects and leverages the softmax to resolve overlap.
For more complex scenes, we also expect iterative refinement to be advantageous for resolving difficult cases. 
There, \modelname could start with a rough segmentation and then use the progressively better understanding of the constituent objects for refining the boundaries.

The segmentation process of MONet is deterministic which induces an order on the objects, which might be useful because it naturally prioritizes salient objects.
We observed that it typically starts with the background, then processes large frontal objects, and finally smaller or farther away objects.
But this approach does break symmetry between objects, and we prefer keeping such a bias out of the object segmentation learning as much as possible.

Another disadvantage of a deterministic segmentation is that it cannot directly deal with ambiguous cases like the one shown in \cref{sec:results:multistability} and \cref{fig:tetris}.
The iterative message-passing-like approach of \modelname might also lend itself well for incorporating top-down feedback to bias the segmentation towards one that is useful for a given task. 
It is less clear how to do that in MONet, though adding a way for conditioning the masking network could potentially serve a similar purpose.
Finally the iterative refinement of \modelname naturally extends to sequential data (see \cref{appendix:discussion:sequences}) which would be less straightforward for MONet.

In summary, it is not at all clear yet which approach will work better and under which circumstances.
If the data is sequential or contains ambiguity, \modelname presents a better choice.
For other data that is not visually more complex than CLEVR, both methods will likely produce similar results making MONet the simpler and less computationally intensive choice.
For more complex data it is unclear yet which approach would be the better choice, and in fact a hybrid approach might be the most promising.
Sequentially attending to objects and iterative refinement are not mutually exclusive and might support each other.
We consider this a very attractive research direction and are excited to explore its possibilities.

\section{Dataset Details}
\label{appendix:data}
\subsection{CLEVR}

We regenerated the CLEVR dataset~\cite{johnson2017clevr} using the authors' open-source code, because we needed ground-truth segmentation masks for evaluation purposes. 
The dataset contains 70\,000 images with a resolution of $240\times320$ pixels, from which we extract a square center crop of $192\times192$ and scale it to $128\times128$ pixels.
Each scene contains between three and ten objects, characterized in terms of shape (cube, cylinder, or sphere), size (small or large), material (rubber or metal), color (8 different colors), position (continuous), and rotation (continuous).

The subset of images which contain 3-6 objects (inclusive) served as the training set for our experiments; we refer to it as CLEVR6. Unless noted otherwise, we evaluate models on the full CLEVR distribution, containing 3-10 objects. All references to CLEVR refer to the full distribution.

\begin{figure}[h!]
    \centering
    \vspace*{-5pt}
    \includegraphics[width=0.95\columnwidth]{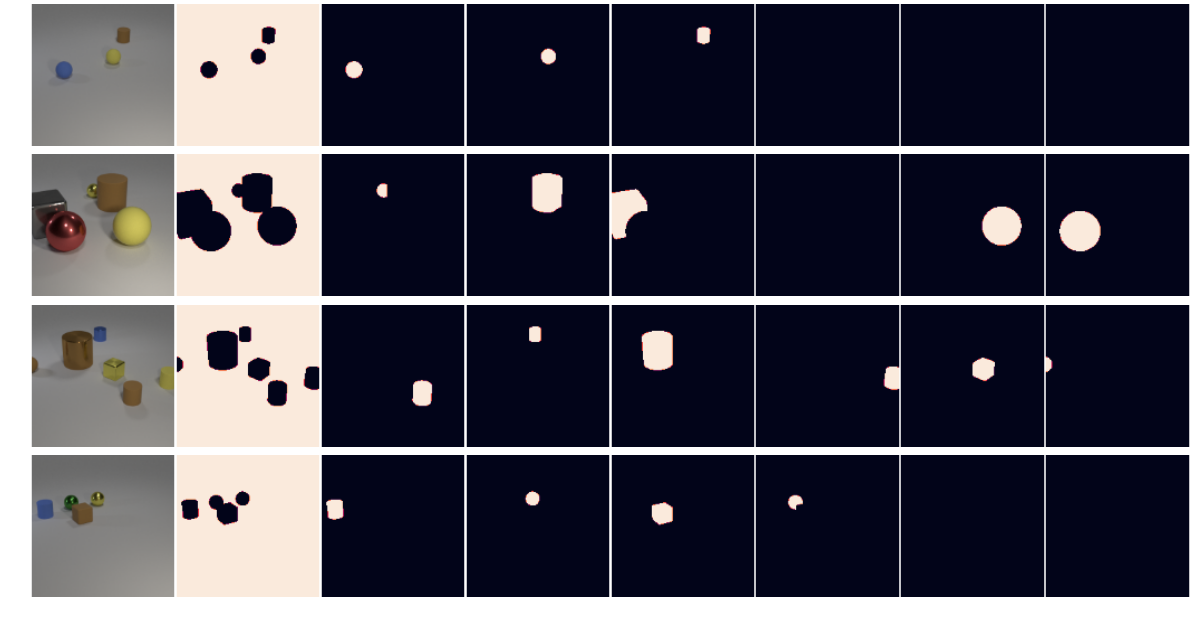}
    \vspace*{-10pt}
    \caption{Samples from CLEVR6. The first column is the scene, the second column is the background mask and the following columns are the ground-truth object masks.}
    \vspace*{-10pt}
    \label{fig:demo_clevr}
\end{figure}

We do not make use of the question answering task. \cref{fig:demo_clevr} shows a few samples from the dataset.

\subsection{Multi-dSprites}

\begin{figure}[b!]
    \centering
    \includegraphics[width=0.95\columnwidth]{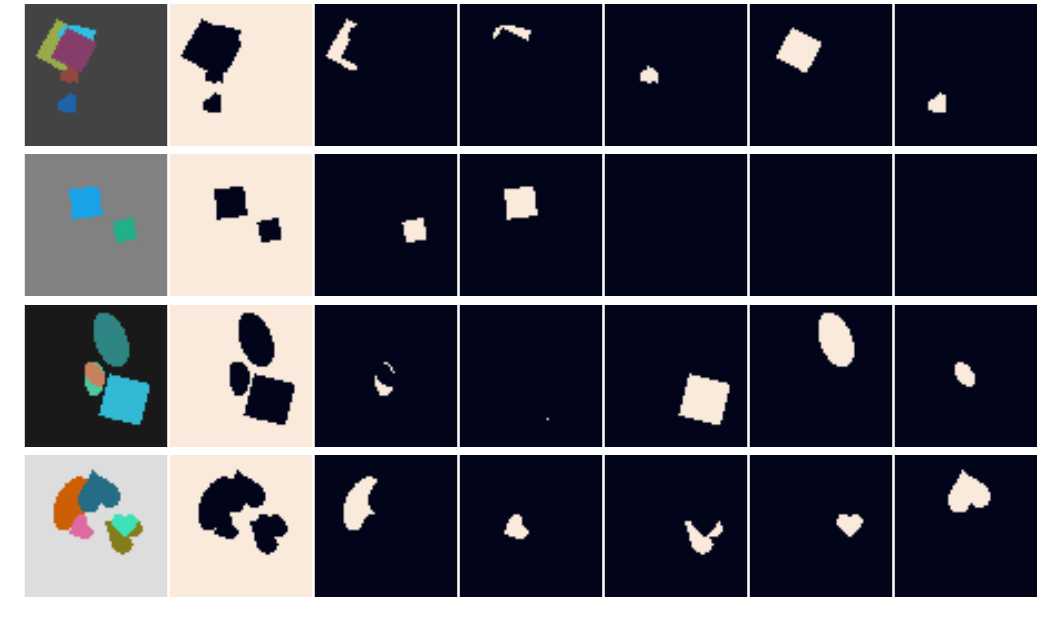}
    \vspace*{-10pt}
    \caption{Samples from the Multi-dSprites dataset. The first column is the full image, the second column is the background mask and the following columns are the ground-truth object masks.}
    \vspace*{-5pt}
    \label{fig:demo_dsprites}
\end{figure}

This dataset, based on the dSprites dataset~\cite{matthey2017dsprites}, consists of 60\,000 images with a resolution of $64\times64$.
Each image contains two to five random sprites, which vary in terms of shape (square, ellipse, or heart), color (uniform saturated colors), scale (continuous), position (continuous), and rotation (continuous).
Furthermore the background color is varied in brightness but always remains grayscale. \cref{fig:demo_dsprites} shows a few samples from the dataset.

We also used a binarized version of Multi-dSprites, where the sprites are always white, the background is always black, and each image contains two to three random sprites.

\subsection{Tetris}

\begin{figure}[t!]
    \centering
    \includegraphics[width=0.95\columnwidth]{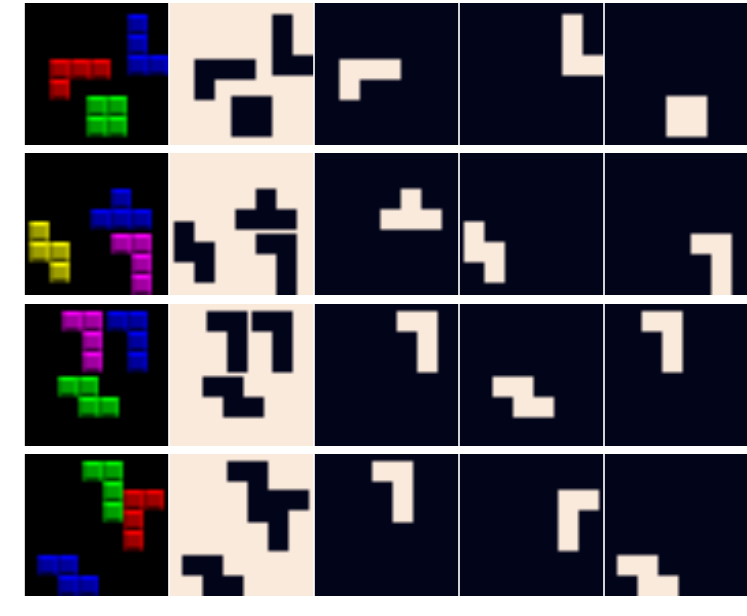}
    \vspace*{-10pt}
    \caption{Samples from the Tetris dataset. The first column is the full image, the second column is the background mask and the following columns are the ground-truth object masks.}
    \label{fig:demo_tetris}
\end{figure}

We generated this dataset of 60\,000 images by placing three random Tetrominoes without overlap in an image of $35\times35$ pixels. 
Each Tetromino is composed of four blocks that are each $5\times5$ pixels.
There are a total of 17 different Tetrominoes (counting rotations).
We randomly color each Tetromino with one of 6 colors (red, green, blue, cyan, magenta, or yellow). \cref{fig:demo_tetris} shows a few samples from the dataset.

\subsection{Shapes}
We use the same shapes dataset as in \cite{reichert2013neuronal}.
It contains 60\,000 binary images of size $28\times28$ each with three random shapes from the set $\{\triangle, \raisebox{0.2em}{\(\bigtriangledown\)}, \square\}$.

\subsection{Objects Room}
For the preliminary sequential experiments we used a sequential version of the \emph{Objects Room} dataset \citep{burgess2019monet}. This dataset consists of 64x64 RGB images of a cubic room, with randomly colored walls, floors and objects randomly scattered around the room. The camera is always positioned on a ring inside the room, always facing towards the centre and oriented vertically in the range $(-25^\circ, 22^\circ)$. There are 3 randomly shaped objects in the room with 1-3 objects visible in any given frame. This version contains sequences of camera-flights for 16 time steps, with the camera position and angle (within the above constraints) changing according to a fixed velocity for the entire sequence (with a random velocity sampled for each sequence).


\section{Model and Hyperparameter Details}
\label{appendix:model}
\paragraph{Training}
Unless otherwise specified all the models are trained with the ADAM optimizer~\cite{kingma2015adam}, with default parameters and a learning rate of $0.0003$. 
We used gradient clipping as recommended by \cite{pascanu2012understanding}: if the norm of global gradient exceeds $5.0$ then the gradient is scaled down to that norm. 
Note that this is virtually always the case as the gradient norm is typically on the order of $10^5$, but we nonetheless found it useful to apply this strategy. 
We always use $\sigma=0.1$ for the global scale of the output distribution $p(\x|\z^{(t)}) = \mathcal{N}(\x; \bm{\mu}^{(t)}_k, \sigma^2)$.
Finally, batch size was 32 ($4 \times 8$GPUs).

\paragraph{Initialization of Posterior}
\modelname iteratively refines an initial posterior $\bm{\lambda}^{(1)}$ which is independent of the input data.
Initially we set this initial value to match the prior (i.e. $q_{\bm{\lambda}}(\z^{(1)}_k) = \mathcal{N}(\bm{0}, \bm{1})$).
But we found that this poses problems for the model, because of the competing requirements it poses for structuring the latent space w.r.t. the prior:
On the one hand, samples from the prior need to be good starting values for iterative refinement.
On the other hand, the prior should correspond to the accumulated posterior (KL term).
For this reason we decided to simply make the parameters $\bm{\lambda}^{(1)}$ of the initialization distribution trainable parameters which are optimized alongside the weights of the decoder ($\bm{\theta}$) and of the refinement network ($\bm{\phi}$).
This lead to faster training, and improved the visual quality of reconstructions from prior samples.

\paragraph{Inputs}
For all models, we use the following inputs to the refinement network, where LN means Layernorm and SG means stop gradients, and we omit the iteration index $\cdot^{(t)}$ for brevity.
The following image-sized inputs are concatenated and fed to the corresponding convolutional network:
\begin{tabular}{llccc}
\toprule
    Description & Formula & LN & SG & Ch.\\ \midrule
    image & $\x$ & & & 3 \\
    means & $\bm{μ}$ &  & & 3  \\
    mask  & $\bm{m}_k$ & & & 1 \\
    mask-logits & $\bm{\hat{m}}_k$ & & & 1 \\
    mask posterior & $\p(\bm{m}_k|\x,\bm{μ})$ & & & 1 \\
    gradient of means & $\nabla_{\bm{μ}_k}\mathcal{L}$ & \checkmark & \checkmark & 3 \\
    gradient of mask & $\nabla_{\bm{m}_k}\mathcal{L}$ & \checkmark & \checkmark & 1 \\
    pixelwise likelihood & $\p(\x|\z)$ & \checkmark & \checkmark & 1 \\
    leave-one-out likelih. & $\p(\x|\z_{i \neq k})$ & \checkmark & \checkmark & 1 \\
    coordinate channels & & & & 2 \\
    \midrule
    \multicolumn{4}{r}{total:} & 17 \\
    \bottomrule
\end{tabular}

The posterior parameters $\bm{\lambda}$ and their gradients are flat vectors, and as such we concatenate them with the output of the convolutional part of the refinement network and use the result as input to the refinement LSTM:
\begin{tabular}{llcc}
\toprule
    Description & Formula & LN & SG\\ \midrule
    gradient of posterior & $\nabla_{\bm{λ}_k}\mathcal{L}$ & \checkmark & \checkmark \\
    posterior & $\bm{λ}_k$ & &\\
    \bottomrule
\end{tabular}

\paragraph{Architecture}
All layers use the ELU~\cite{clevert2015fast} activation function and the Convolutional layers use a stride equal to 1, unless mentioned otherwise.
Architecture details for the individual datasets are summarized in the following subsections.

\subsection{CLEVR}
All models were trained on scenes with 3-6 objects (CLEVR6) with $K=7$ slots and $T=5$ iterations, and a latent object dimension of size $\text{dim}(\z_k)=64$.
Training was done on eight V100 GPUs for 1M updates (approx. 1 week wall time).
When evaluating on the full CLEVR dataset, we increased the number of slots to $K=11$. For some of the analysis, we varied $T$ and $K$ as mentioned in the text. 

The rest of the architecture and hyperparameters are described in the following.

\begin{tabular}{lccl}
    \multicolumn{4}{l}{\textbf{Decoder}}\\
    \toprule
    Type & Size/\textit{Ch.} & Act. Func. & Comment\\ \midrule
    Input: $\z$  & 64 &  & \\
    Broadcast & \textit{66} & & + coordinates\\
    Conv $3\times3$ & \textit{64} & ELU & \\
    Conv $3\times3$ & \textit{64} & ELU & \\
    Conv $3\times3$ & \textit{64} & ELU & \\
    Conv $3\times3$ & \textit{64} & ELU & \\
    Conv $3\times3$ & \textit{4} & Linear  & RGB + Mask\\
    \bottomrule
\end{tabular}

\begin{tabular}{lccl}
    \multicolumn{4}{l}{\textbf{Refinement Network}}\\
    \toprule
    Type & Size/\textit{Ch.} & Act. Func. & Comment\\ \midrule
    MLP & 128 & Linear & \\
    LSTM & 256 & Tanh & \\
    Concat $[\bm{\lambda}, \nabla_{\bm{\lambda}}\mathcal{L}]$ & 512 & & \\
    MLP & 256 & ELU & \\
    Avg. Pool & 64 & \\
    Conv $3\times3$ & \textit{64} & ELU & stride 2\\
    Conv $3\times3$ & \textit{64}  & ELU & stride 2\\
    Conv $3\times3$ & \textit{64}  & ELU & stride 2\\
    Conv $3\times3$ & \textit{64}  & ELU & stride 2\\
    Inputs & \textit{17}  & & \\
    \bottomrule
\end{tabular}

\begin{tabular}{lccl}
    \multicolumn{4}{l}{\textbf{Deconv Decoder} used in \cref{sec:results:ablation}}\\
    \toprule
    Type & Size/\textit{Ch.} & Act. Func. & Comment\\ \midrule
    Input: $\z$  & 64 &  & \\
    MLP & 512 & ELU & \\
    MLP & 512 & ELU & \\
    Reshape & \textit{8} & & $8\times 8 \times 8$\\
    Conv $5\times5$ & \textit{64} & ELU & stride 2\\
    Conv $5\times5$ & \textit{64} & ELU & stride 2\\
    Conv $5\times5$ & \textit{64} & ELU & stride 2\\
    Conv $5\times5$ & \textit{64} & ELU & stride 2\\
    Conv $5\times5$ & \textit{64} & ELU & \\
    Conv $5\times5$ & \textit{4} & Linear  & RGB + Mask\\
    \bottomrule
\end{tabular}

\subsection{Multi-dSprites}
Models were trained with $K=6$ slots, and used $T=5$ iterations.

\begin{tabular}{lccl}
    \multicolumn{4}{l}{\textbf{Decoder}}\\
    \toprule
    Type & Size/\textit{Ch.} & Act. Func. & Comment\\ \midrule
    Input: $\bm{λ}$  & 32 &  & \\
    Broadcast & \textit{34} & & + coordinates\\
    Conv $5\times5$ & \textit{32} & ELU & \\
    Conv $5\times5$ & \textit{32} & ELU & \\
    Conv $5\times5$ & \textit{32} & ELU & \\
    Conv $5\times5$ & \textit{32} & ELU & \\
    Conv $5\times5$ & \textit{4} & Linear  & RGB + Mask\\
    \bottomrule
\end{tabular}

\begin{tabular}{lccl}
    \multicolumn{4}{l}{\textbf{Refinement Network}}\\
    \toprule
    Type & Size/\textit{Ch.} & Act. Func. & Comment\\ \midrule
    MLP & 32 & Linear & \\
    LSTM & 128 & Tanh & \\
    Concat $[\bm{\lambda}, \nabla_{\bm{\lambda}}\mathcal{L}]$ & 192 & & \\
    MLP & 128 & ELU & \\
    Avg. Pool & 32 & \\
    Conv $5\times5$ & \textit{32}  & ELU & \\
    Conv $5\times5$ & \textit{32}  & ELU & \\
    Conv $5\times5$ & \textit{32}  & ELU & \\
    Inputs & \textit{17}  & & \\
    \bottomrule
\end{tabular}

\subsection{Tetris}
Models were trained with $K=4$ slots, and used $T=5$ iterations.
For Tetris, in contrast to the other models, we did not use an LSTM in the refinement network. 

\begin{tabular}{lccl}
    \multicolumn{4}{l}{\textbf{Decoder}}\\
    \toprule
    Type & Size/\textit{Ch.} & Act. Func. & Comment\\ \midrule
    Input: $\bm{λ}$  & 64 &  & \\
    Broadcast & \textit{66} & & + coordinates\\
    Conv $5\times5$ & \textit{32} & ELU & \\
    Conv $5\times5$ & \textit{32} & ELU & \\
    Conv $5\times5$ & \textit{32} & ELU & \\
    Conv $5\times5$ & \textit{32} & ELU & \\
    Conv $5\times5$ & \textit{4} & Linear  & RGB + Mask\\
    \bottomrule
\end{tabular}

\begin{tabular}{lccl}
    \multicolumn{4}{l}{\textbf{Refinement Network}}\\
    \toprule
    Type & Size/\textit{Ch.} & Act. Func. & Comment\\ \midrule
    MLP & 64 & Linear & \\
    Concat $[\bm{\lambda}, \nabla_{\bm{\lambda}}\mathcal{L}]$ & 256 & & \\
    MLP & 128 & ELU & \\
    Avg. Pool & 32 & \\
    Conv $5\times5$ & \textit{32}  & ELU & \\
    Conv $5\times5$ & \textit{32}  & ELU & \\
    Conv $5\times5$ & \textit{32}  & ELU & \\
    Inputs & \textit{17}  & & \\
    \bottomrule
\end{tabular}

\pagebreak

\section{Additional Plots}
\label{appendix:plots}

\begin{figure*}[t!]
    \centering
    \includegraphics[height=0.45\textheight]{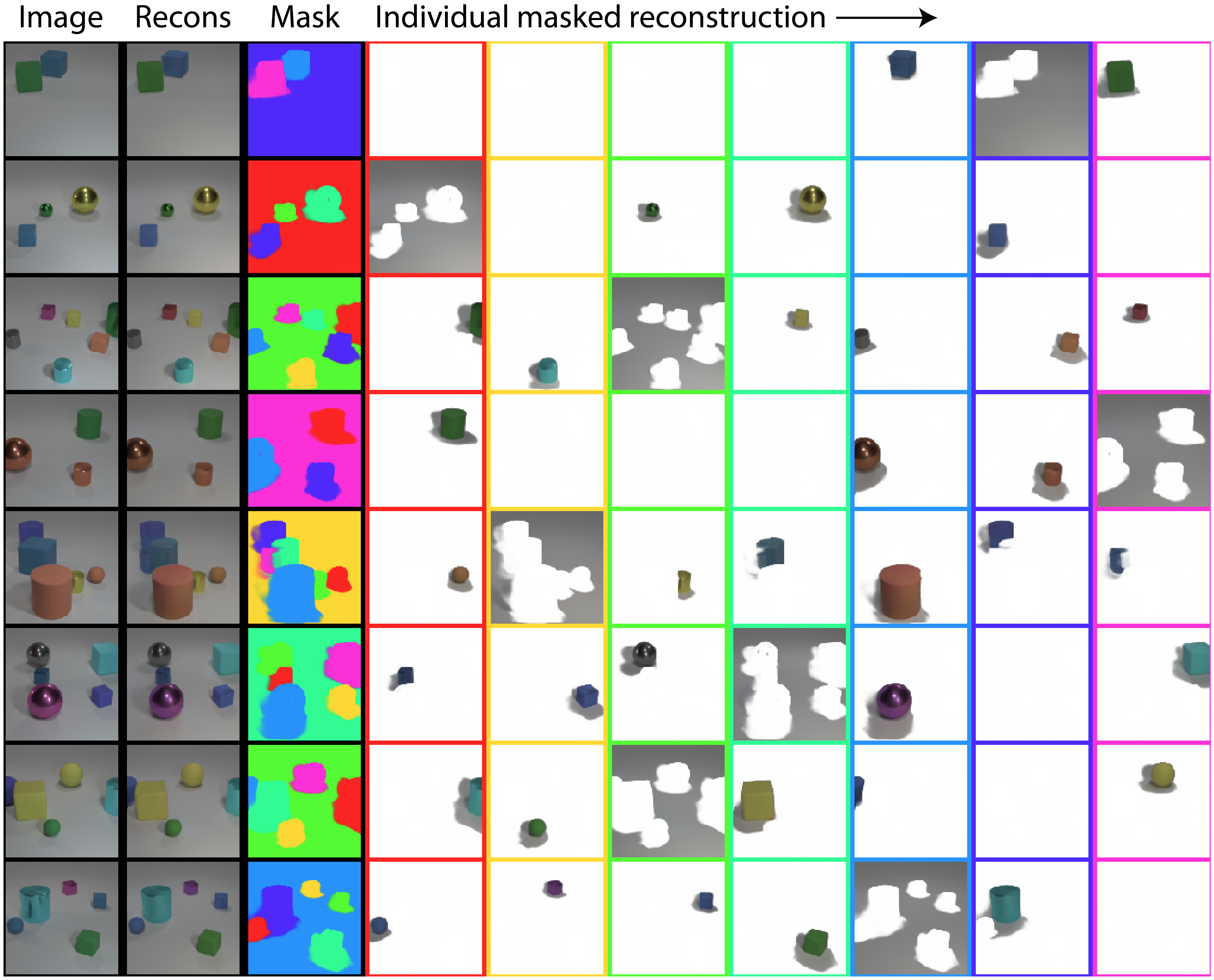}
    \vspace*{-5pt}
    \caption{Additional segmentation and object reconstruction results on CLEVR6. Border colors are matched to the segmentation mask on the left.}
    \vspace*{-5pt}
    \label{fig:additional_clevr}
\end{figure*}

\begin{figure*}[t!]
    \centering
    \vspace*{-5pt}
    \includegraphics[height=0.45\textheight]{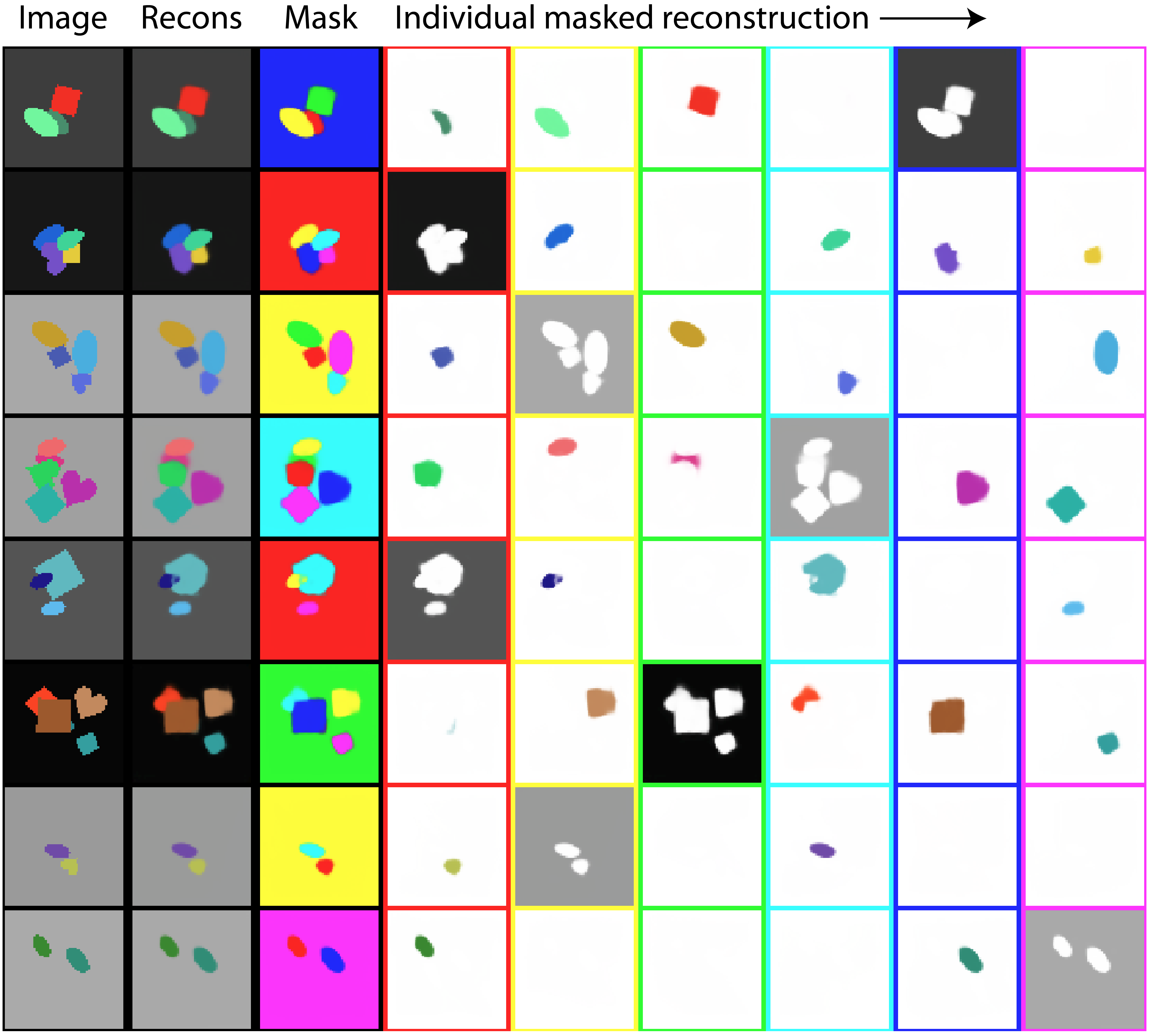}
    \vspace*{-10pt}
    \caption{Additional segmentation and object reconstruction results on Multi-dSprites. Border colors are matched to the segmentation mask on the left.}
    \vspace*{-5pt}
    \label{fig:additional_dsprites}
\end{figure*}

\begin{figure*}[!t]
    \centering
    \vspace*{-5pt}
    \includegraphics[height=0.55\textheight]{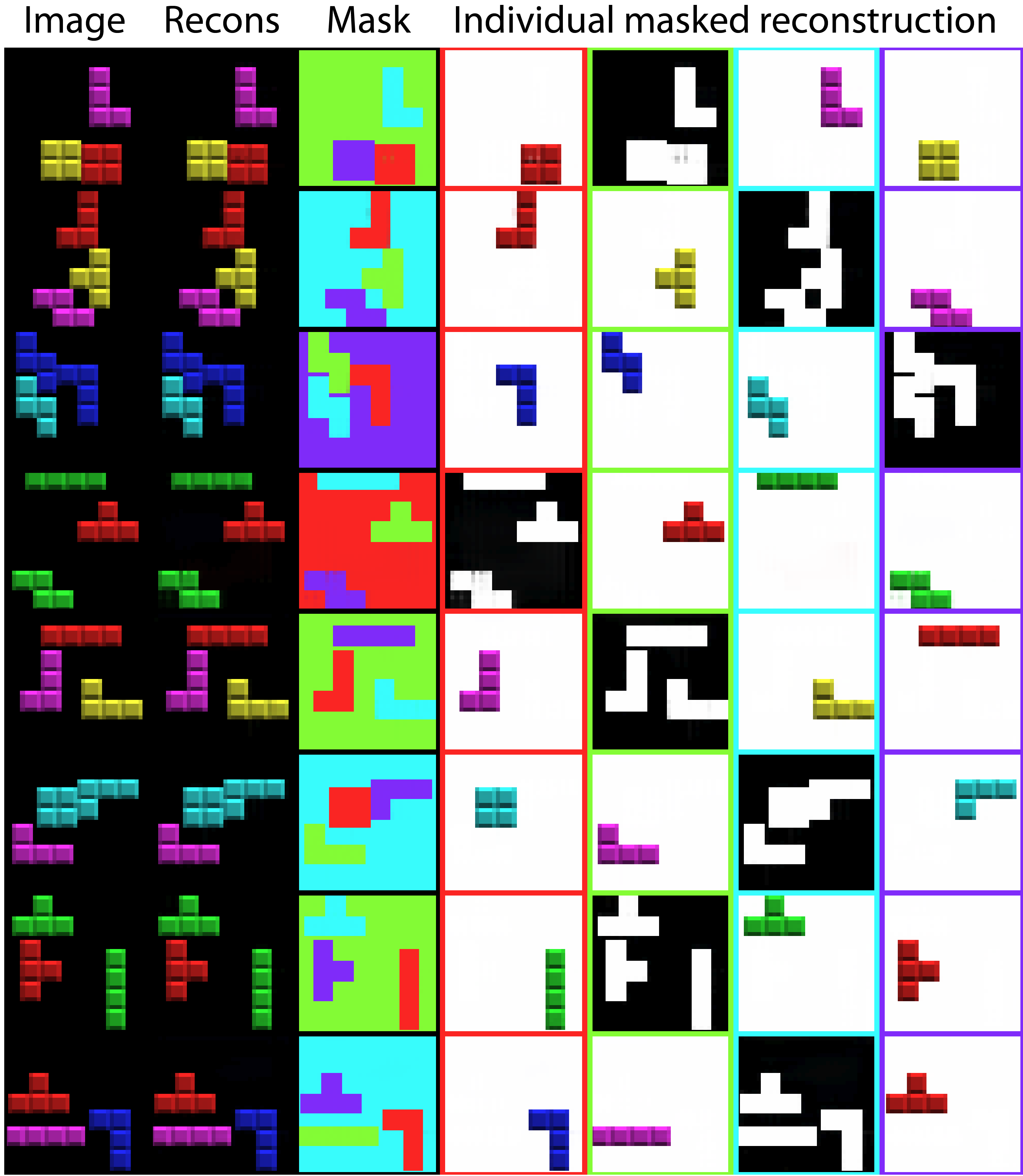}
    \vspace*{-10pt}
    \caption{Additional segmentation and object reconstruction results on Tetris. Border colors are matched to the segmentation mask on the left.}
    \vspace*{-5pt}
    \label{fig:additional_tetris}
\end{figure*}

\begin{figure*}[!t]
    \centering
    \vspace*{-10pt}
    \includegraphics[width=\textwidth]{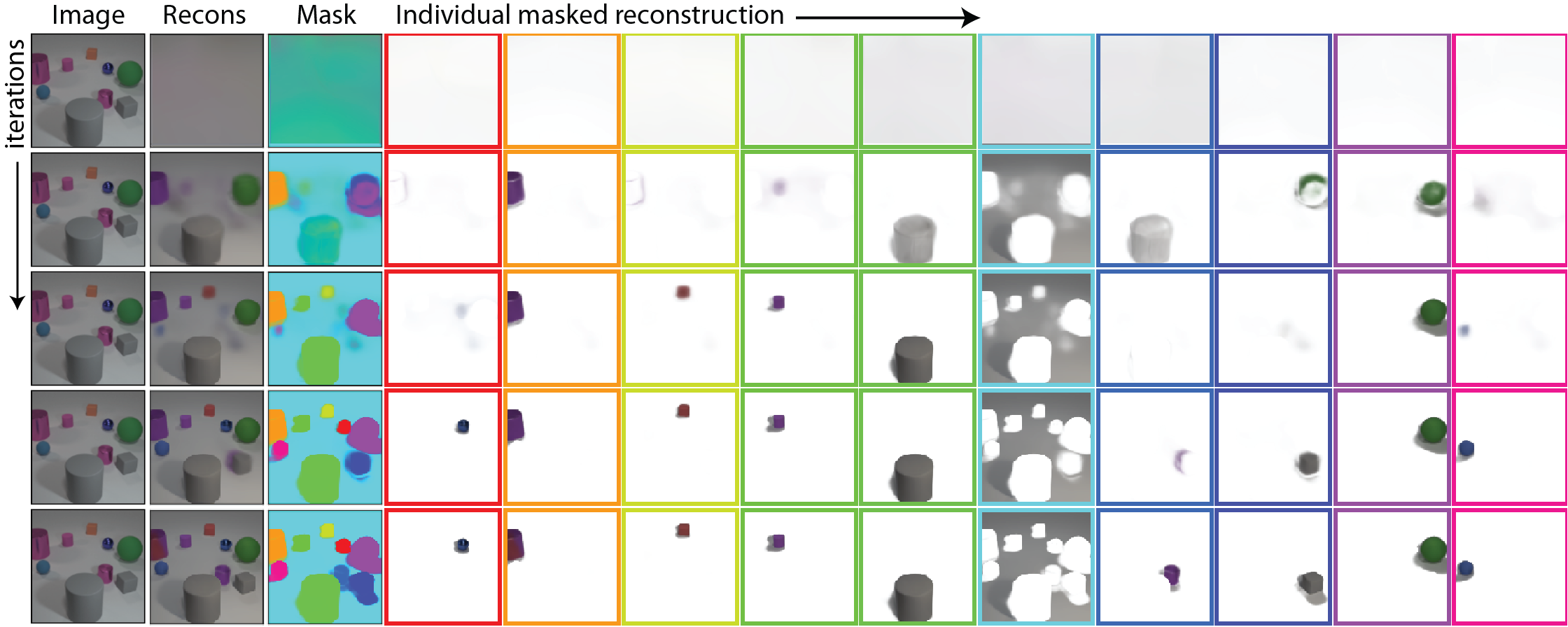}
    \vspace*{-10pt}
    \caption{Full version of Figure~\ref{fig:generalize_color}, showcasing all slots.
    }
    \vspace*{-10pt}
    \label{fig:generalize_color_full}
\end{figure*}

\begin{figure}[t!]
    \centering
    \includegraphics[height=0.62\textheight]{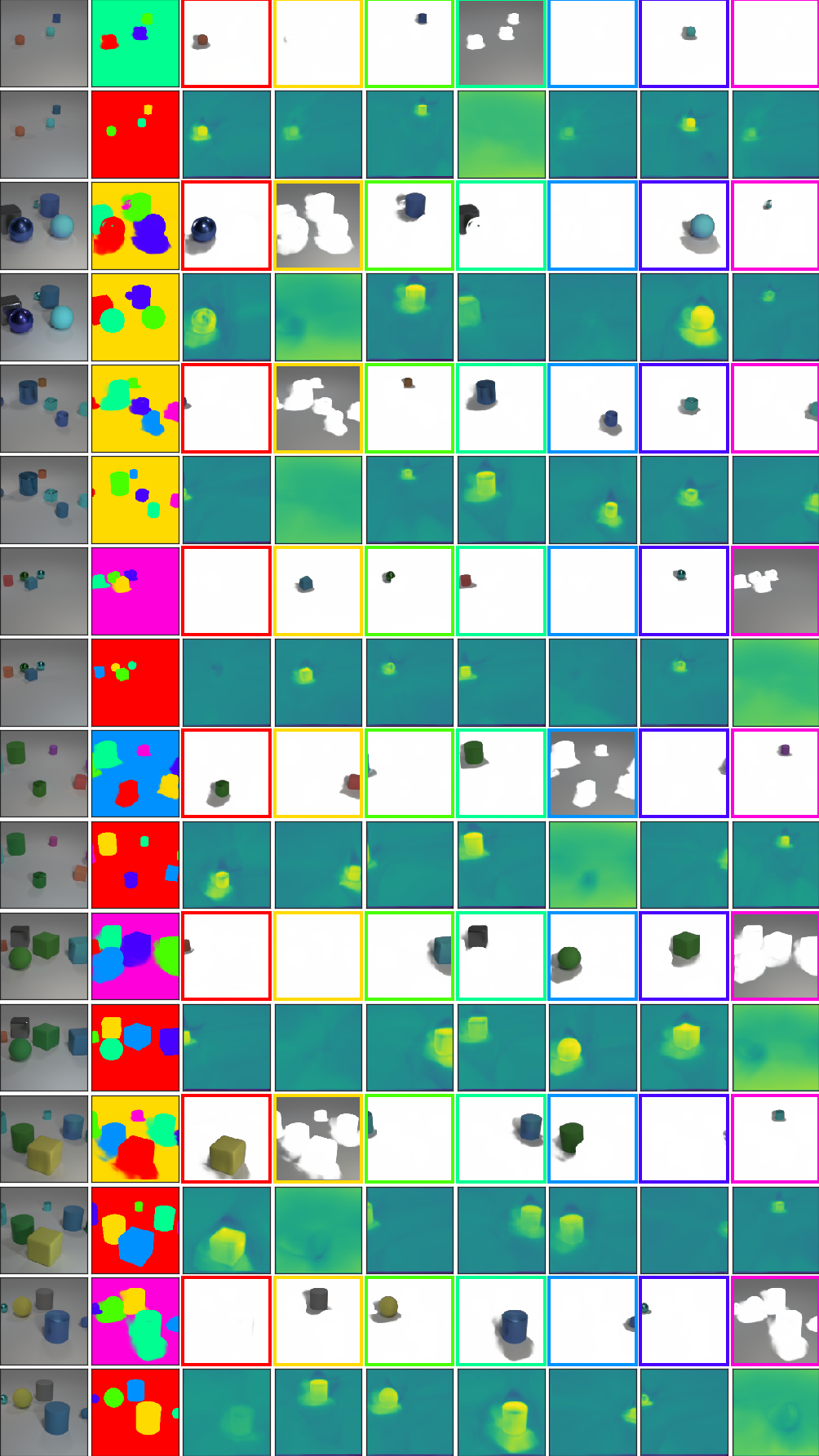}
    \vspace*{-5pt}
    \caption{\textbf{CLEVR6} dataset. Odd rows: image and object masks as determined by the model. Even rows: first column is the input image, second one is the ground-truth masks and the following ones are mask logits produced by the model.}
    \vspace*{-5pt}
    \label{fig:logits_clevr}
\end{figure}

\begin{figure}[t!]
    \centering
    \includegraphics[height=0.62\textheight]{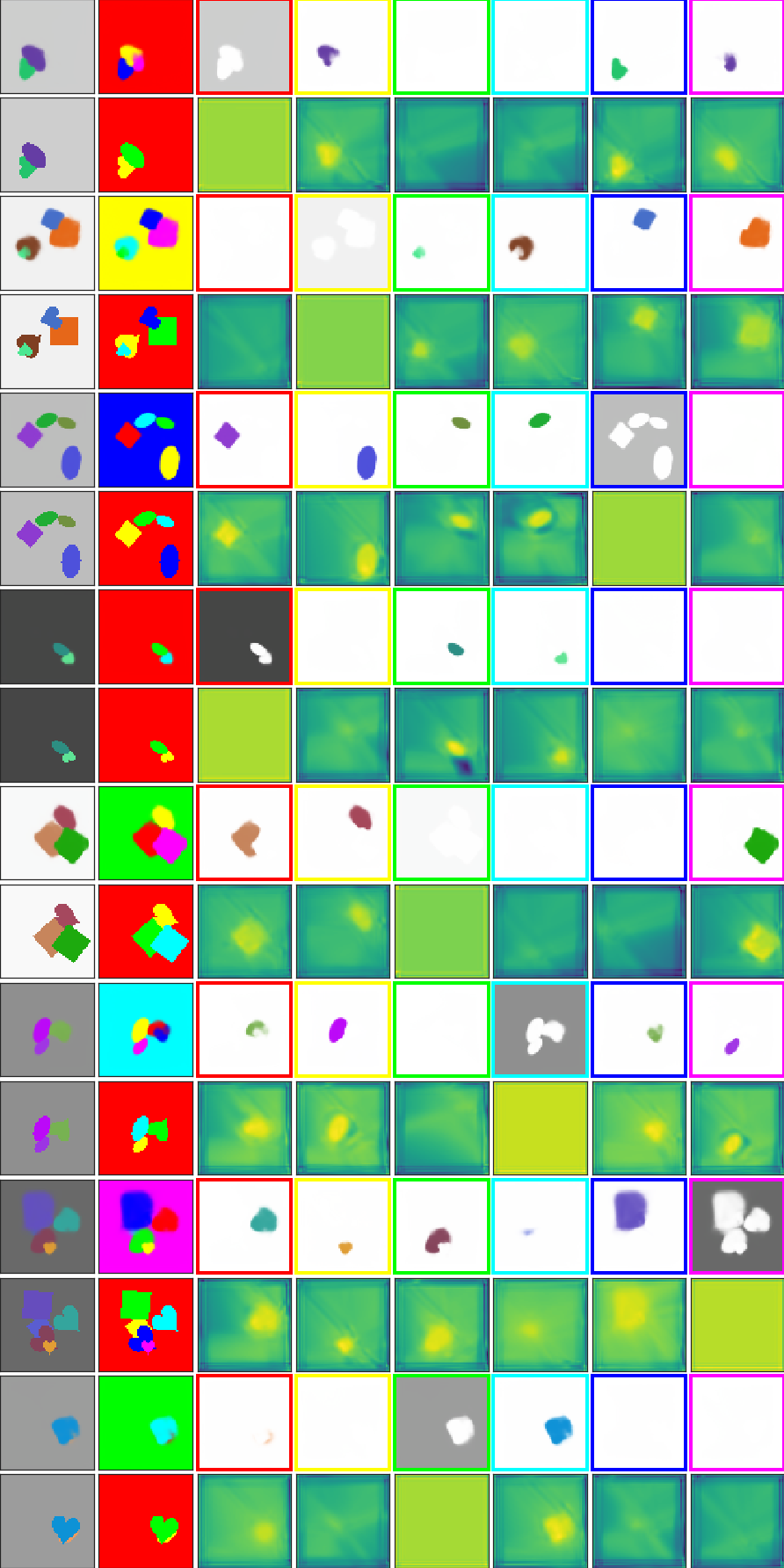}
    \vspace*{-5pt}
    \caption{\textbf{Multi-dSprites} dataset. Odd rows: image and object masks as determined by the model. Even rows: first column is the input image, second one is the ground-truth masks and the following ones are mask logits produced by the model.}
    \vspace*{-5pt}
    \label{fig:logits_sprites}
\end{figure}

\begin{figure*}[t!]
    \centering
    \includegraphics[height=0.9\textheight]{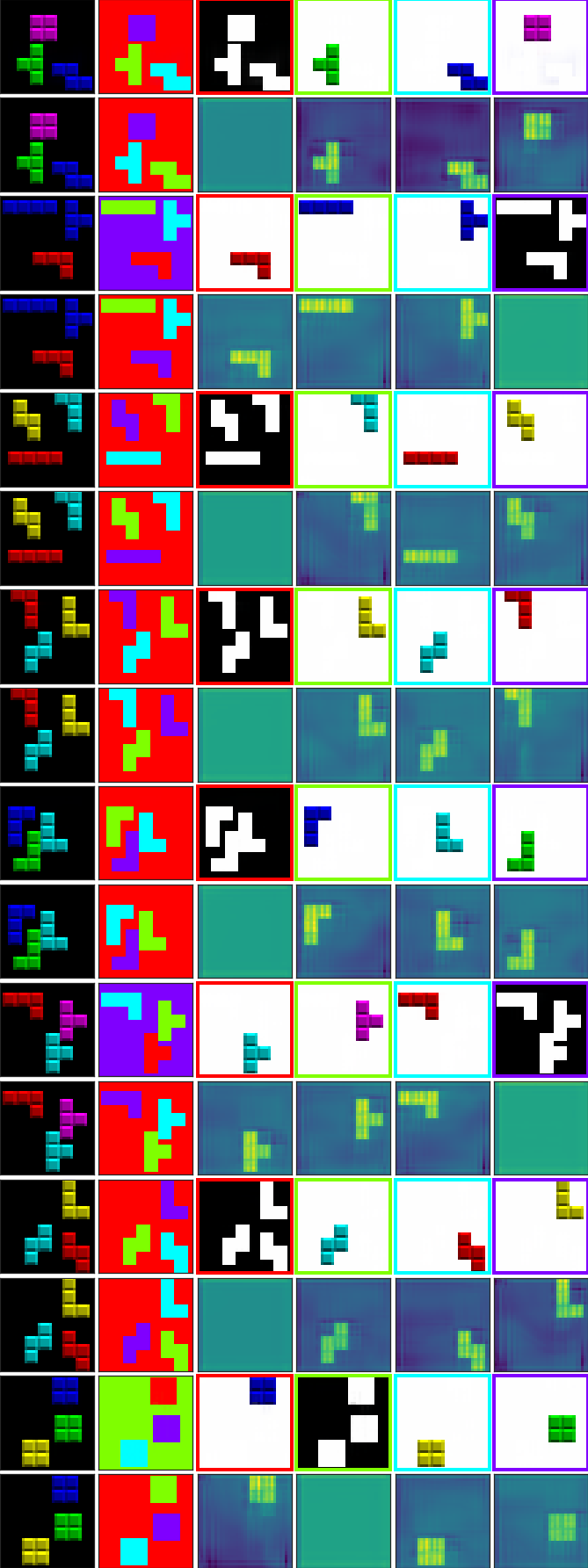}
    \vspace*{-5pt}
    \caption{\textbf{Tetris} dataset. Odd rows: image and object masks as determined by the model. Even rows: first column is the input image, second one is the ground-truth masks and the following ones are mask logits produced by the model.}
    \vspace*{-5pt}
    \label{fig:logits_tetris}
\end{figure*}

\paragraph{Decompositions}
Figures~\ref{fig:additional_clevr}--\ref{fig:additional_tetris} show additional decomposition samples on our datasets.
Figure~\ref{fig:generalize_color_full} shows a complete version of Figure~\ref{fig:generalize_color}, showing all individual masked reconstruction slots. 
Figures~\ref{fig:logits_clevr}--\ref{fig:logits_tetris} show a comparison between the object reconstructions and the mask logits used for assigning decoded latents to pixels.

\begin{figure*}[t!]
    \centering
    \includegraphics[width=0.9\textwidth]{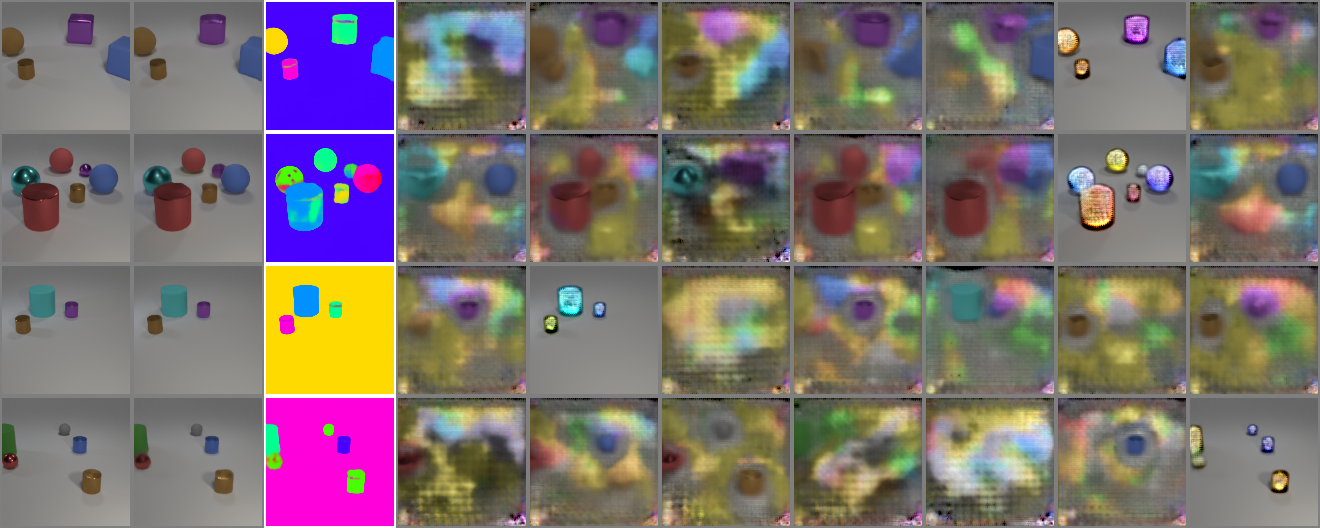}
    \includegraphics[width=0.9\textwidth]{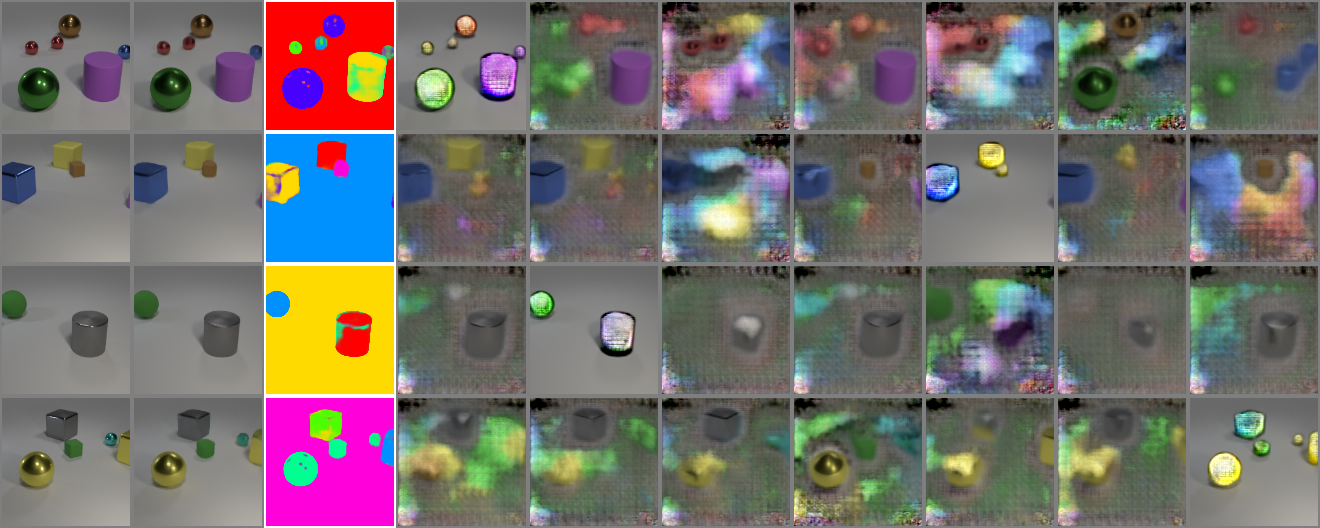}
    \vspace*{-5pt}
    \caption{Segmentation and object reconstruction results on CLEVR6 using a deconvolution based decoder instead of the spatial broadcast decoder.
    Note that \modelname still cleanly segments objects from the background (now ignoring shadows), but specialization of the individual slots is much worse.
    Both, slots holding multiple objects, and objects replicated across multiple slots are much more frequent now.
    Slot reconstructions are also much less clean, containing much more noise and residue of other objects.
    (Note though, that in this figure we didn't mask the reconstructions as we have for \cref{fig:additional_clevr}.)}
    \vspace*{-5pt}
    \label{fig:broadcast_ablation}
\end{figure*}



\begin{figure*}[t!]
    \centering
    \includegraphics[width=0.92\textwidth]{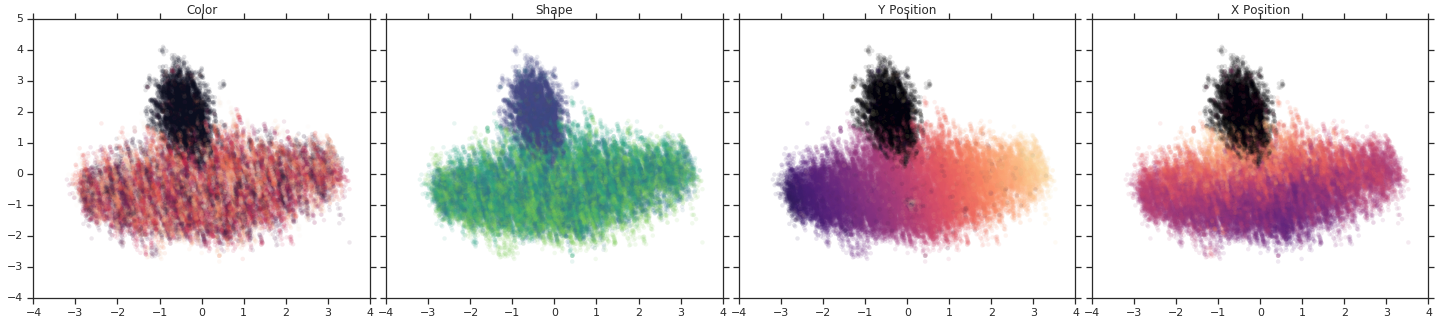}
    \vspace*{-5pt}
    \caption{Projection on the first two principal components of the latent distribution for the \textbf{CLEVR6} dataset. Each dot represents one object latent and is colored according to the corresponding ground truth factor.}
    \vspace*{-5pt}
    \label{fig:pca_clevr}
\end{figure*}
\begin{figure*}[t!]
    \centering
    \includegraphics[width=0.92\textwidth]{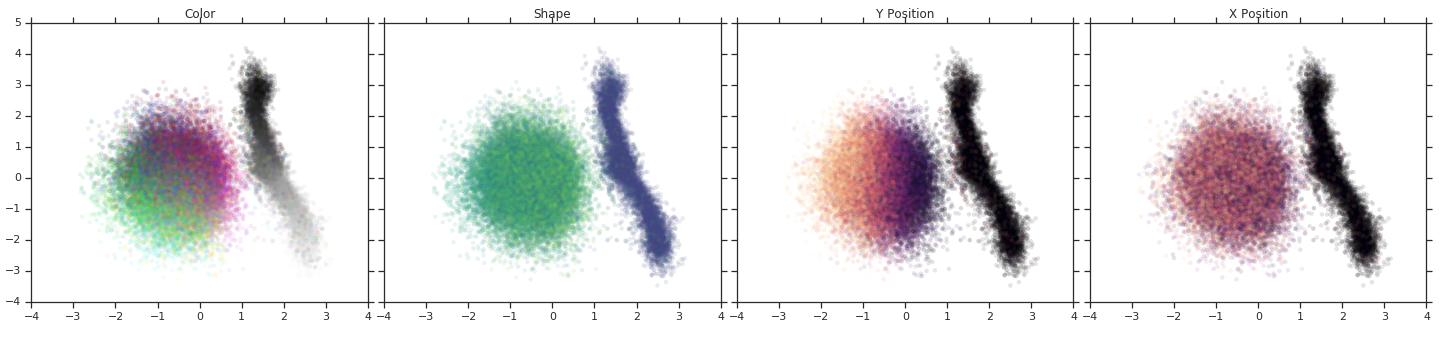}
    \vspace*{-5pt}
    \caption{Projection on the first two principal components of the latent distribution for the \textbf{Multi-dSprites} dataset. Each dot represents one object latent and is colored according to the corresponding ground truth factor.}
    \vspace*{-5pt}
    \label{fig:pca_sprites}
\end{figure*}
\begin{figure*}[t!]
    \centering
    \includegraphics[width=0.92\textwidth]{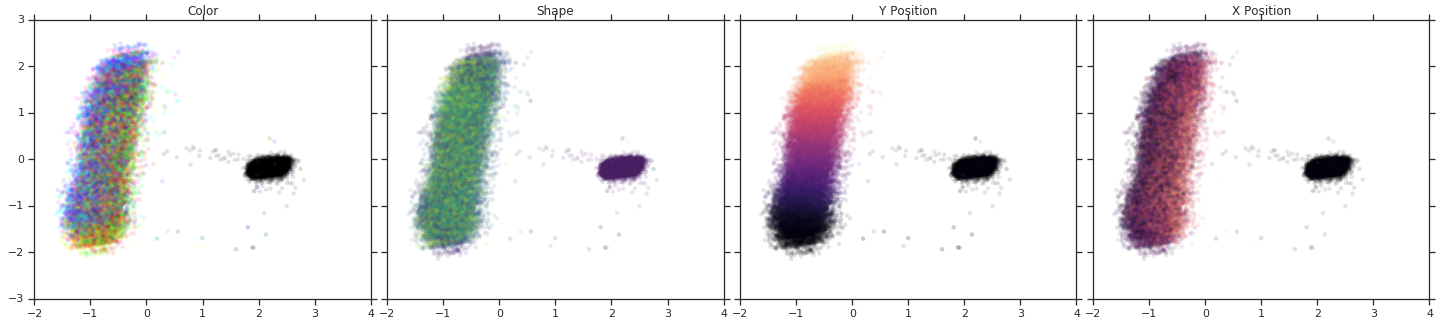}
    \vspace*{-5pt}
    \caption{Projection on the first two principal components of the latent distribution for the \textbf{Tetris} dataset. Each dot represents one object latent and is colored according to the corresponding ground truth factor.}
    \vspace*{-5pt}
    \label{fig:pca_tetris}
\end{figure*}
\begin{figure*}[t!]
    \centering
    \includegraphics[width=0.92\textwidth]{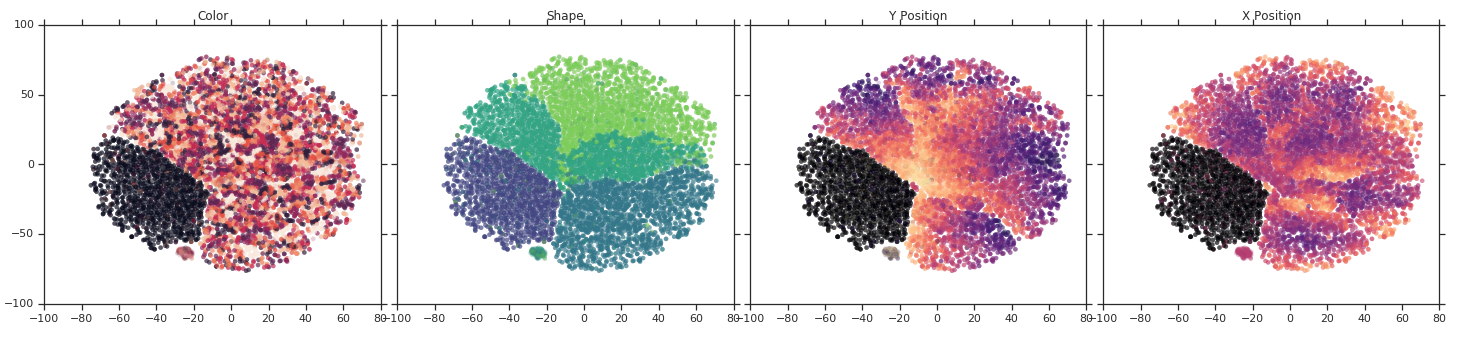}
    \vspace*{-5pt}
    \caption{t-SNE of the latent distribution for the \textbf{CLEVR6} dataset. Each dot represents one object latent and is colored according to the corresponding ground truth factor}
    \vspace*{-5pt}
    \label{fig:tsne_clevr}
\end{figure*}
\begin{figure*}[t!]
    \centering
    \includegraphics[width=0.92\textwidth]{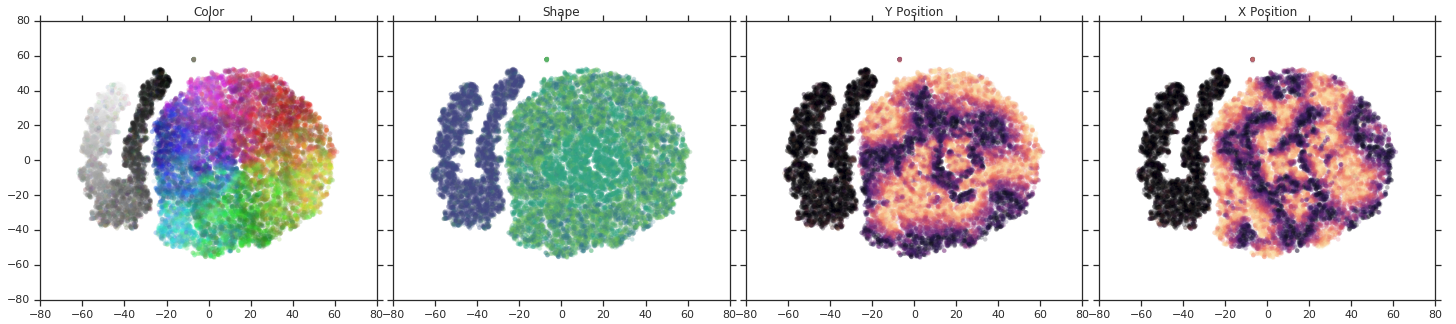}
    \vspace*{-5pt}
    \caption{t-SNE of the latent distribution for the \textbf{Multi-dSprites} dataset. Each dot represents one object latent and is colored according to the corresponding ground truth factor}
    \vspace*{-5pt}
    \label{fig:tsne_sprites}
\end{figure*}
\begin{figure*}[t!]
    \centering
    \includegraphics[width=0.92\textwidth]{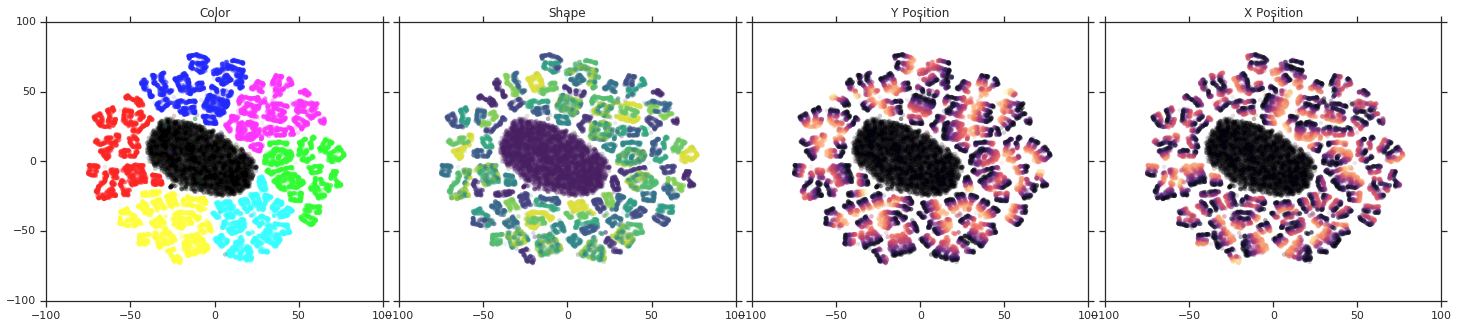}
    \vspace*{-5pt}
    \caption{t-SNE of the latent distribution for the \textbf{Tetris} dataset. Each dot represents one object latent and is colored according to the corresponding ground truth factor}
    \vspace*{-5pt}
    \label{fig:tsne_tetris}
\end{figure*}

\paragraph{Projections of Object Latents}
Figures~\ref{fig:pca_clevr}--\ref{fig:pca_tetris} demonstrate how object latents are clustered when projected onto the first two principal components of the latent distribution. 
Figures~\ref{fig:tsne_clevr}--\ref{fig:tsne_tetris} show how object latents are clustered when projected onto a t-SNE \citep{maaten2008visualizing} of the latent distribution. 

\begin{figure*}[t!]
    \centering
    \includegraphics[width=0.92\textwidth]{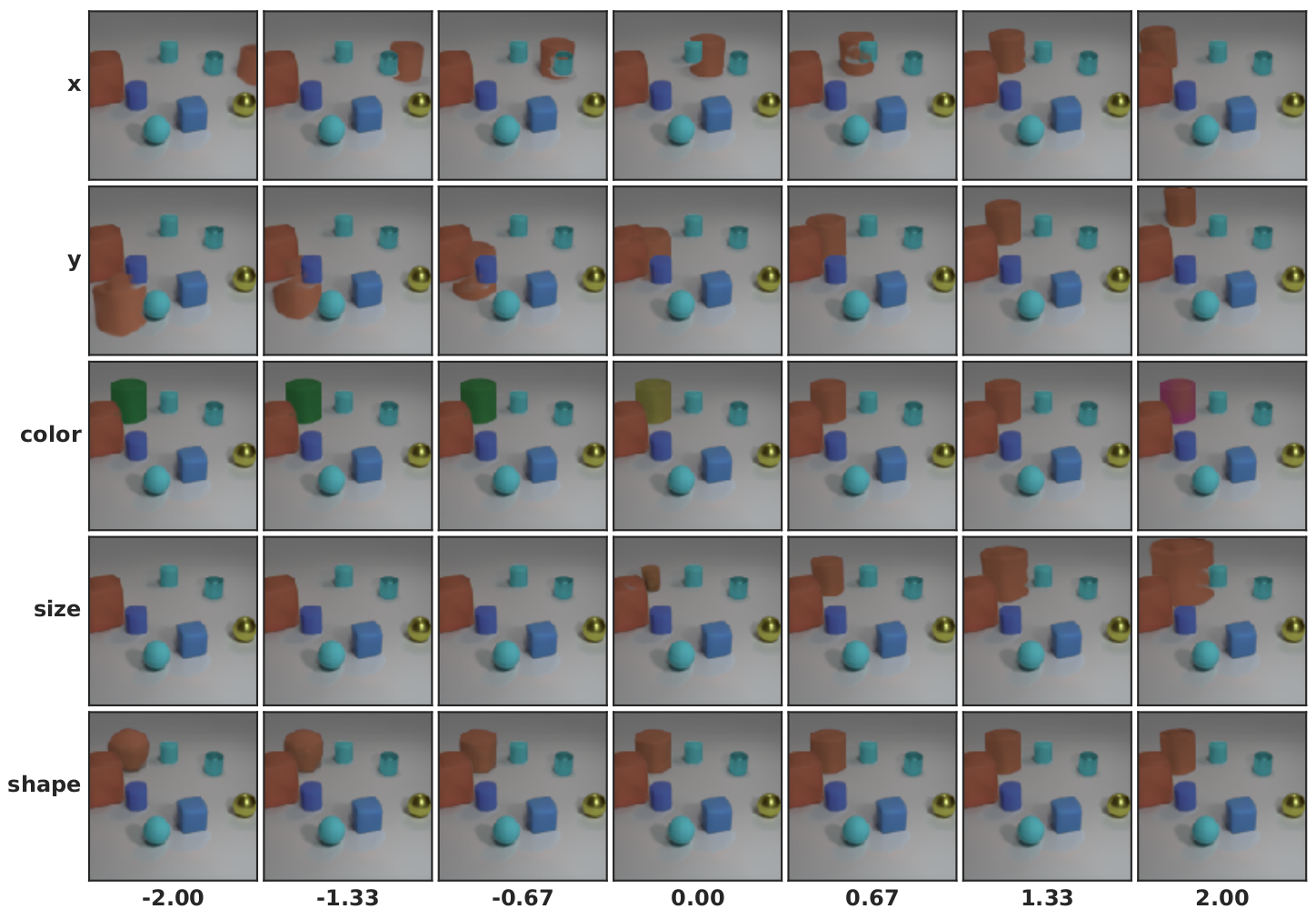}
    \vspace*{-5pt}
    \caption{Latent traversal of IODINE on CLEVR (like right side of \cref{fig:traversals}), for a randomly chosen example and randomly chosen slot. Here the brown cylinder in the back is changing. Occlusion handling shows several flaws, that could be fixed by adjusting another latent (not shown) that encodes the depth ordering.}
    \vspace*{-5pt}
    \label{fig:bonus_traversal1}
\end{figure*}

\begin{figure*}[t!]
    \centering
    \includegraphics[width=0.92\textwidth]{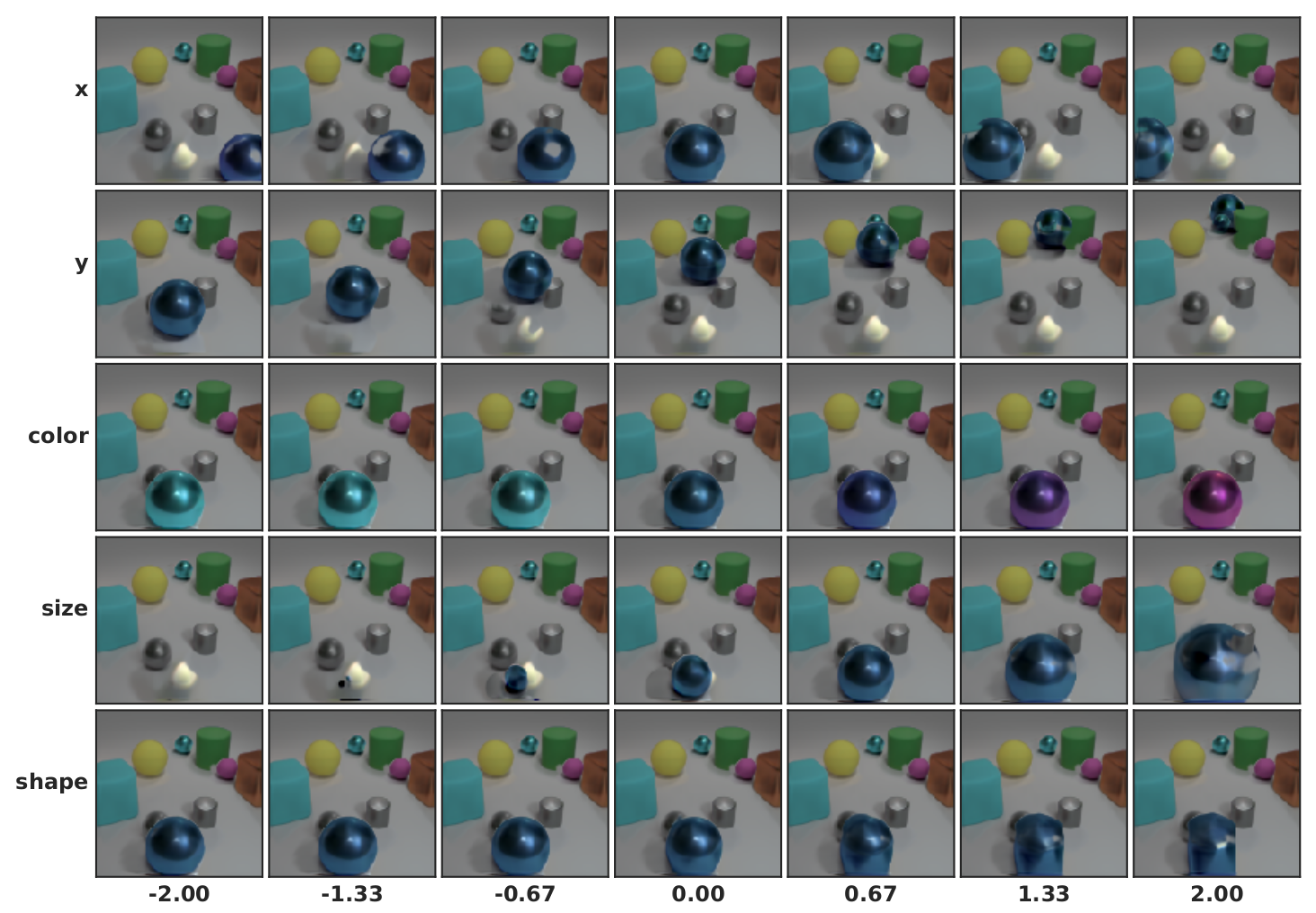}
    \vspace*{-5pt}
    \caption{Latent traversal of IODINE on CLEVR (like right side of \cref{fig:traversals}), for a randomly chosen example and randomly chosen slot. Here the large blue sphere in the front is changing. Note that the background slot contains a bright spot behind the blue sphere that becomes visible when the sphere is moved away. }
    \vspace*{-5pt}
    \label{fig:bonus_traversal2}
\end{figure*}

\begin{figure*}[t!]
    \centering
    \includegraphics[width=0.92\textwidth]{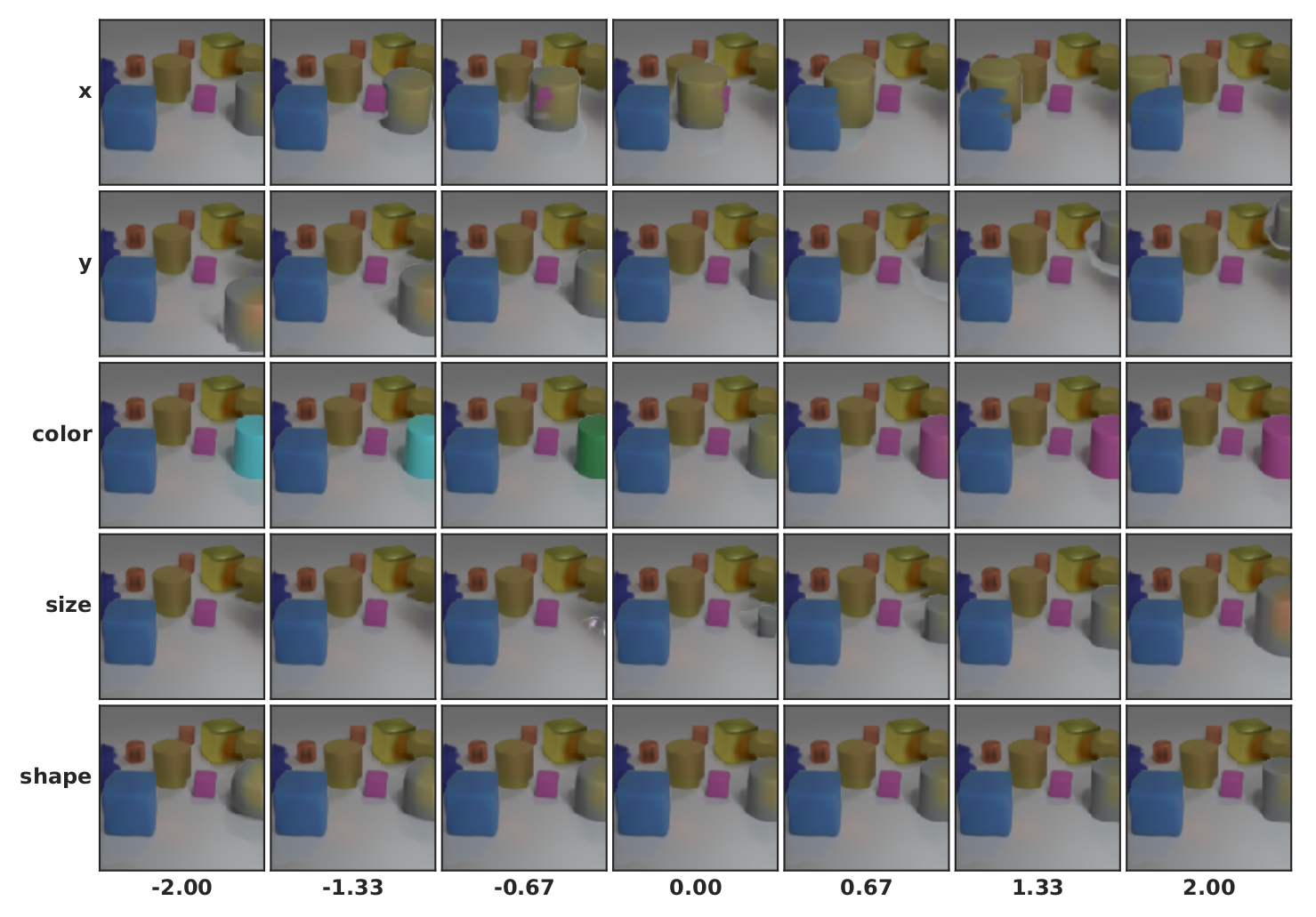}
    \vspace*{-5pt}
    \caption{Latent traversal of IODINE on CLEVR (like right side of \cref{fig:traversals}), for a randomly chosen example and randomly chosen slot. Here the gray cylinder on the right is changing. Occlusion handling shows several flaws, that could be fixed by adjusting another latent (not shown) that encodes the depth ordering.}
    \vspace*{-5pt}
    \label{fig:bonus_traversal3}
\end{figure*}

\paragraph{Traversals}
Figures~\ref{fig:bonus_traversal1}--\ref{fig:bonus_traversal3} show additional (randomly chosen) latent traversals for \modelname on CLEVR like on the right side of \cref{fig:traversals}.

\begin{figure*}[t!]
    \centering
    \includegraphics[width=0.92\textwidth]{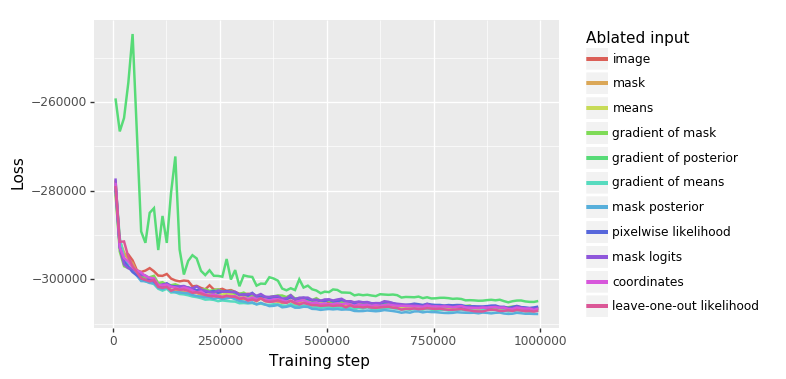}
    \caption{Ablation study for the model's total loss on CLEVR6. Each curve denotes the result of training the model without a particular input.}
    \vspace*{-5pt}
    \label{fig:abl_clevr_loss}
\end{figure*}

\begin{figure*}[t!]
    \centering
    \includegraphics[width=0.92\textwidth]{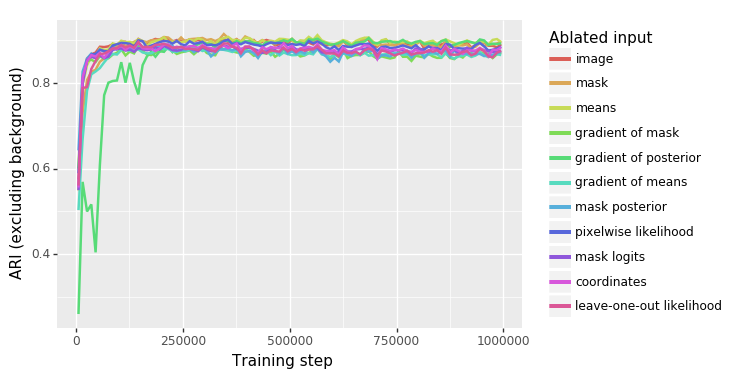}
    \caption{Ablation study for the model's segmentation performance in terms of ARI (excluding the background pixels) on CLEVR6. Each curve denotes the result of training the model without a particular input.}
    \vspace*{-5pt}
    \label{fig:abl_clevr_ari}
\end{figure*}

\begin{figure*}[t!]
    \centering
    \includegraphics[width=0.92\textwidth]{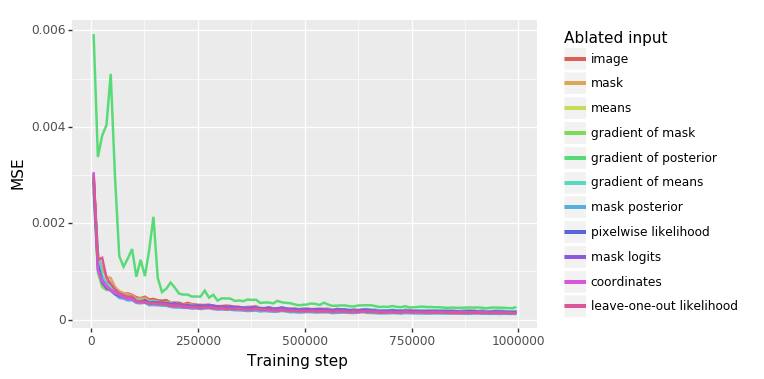}
    \caption{Ablation study for the model's reconstruction loss term on CLEVR6. Each curve denotes the result of training the model without a particular input. The y-axis shows the mean squared error between the target image and the output means (of the final iteration) as a proxy for the full reconstruction loss.}
    \vspace*{-5pt}
    \label{fig:abl_clevr_mse}
\end{figure*}

\begin{figure*}[t!]
    \centering
    \includegraphics[width=0.92\textwidth]{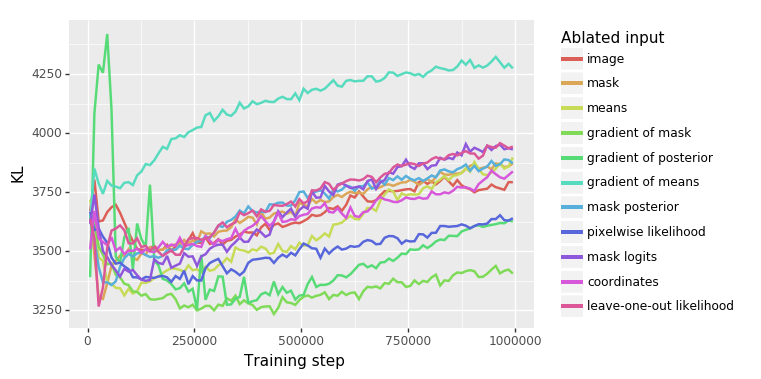}
    \caption{Ablation study for the model's KL divergence loss term on CLEVR6, summed over slots and iterations. Each curve denotes the result of training the model without a particular input.}
    \vspace*{-5pt}
    \label{fig:abl_clevr_kl}
\end{figure*}

\begin{figure*}[t!]
    \centering
    \includegraphics[width=0.92\textwidth]{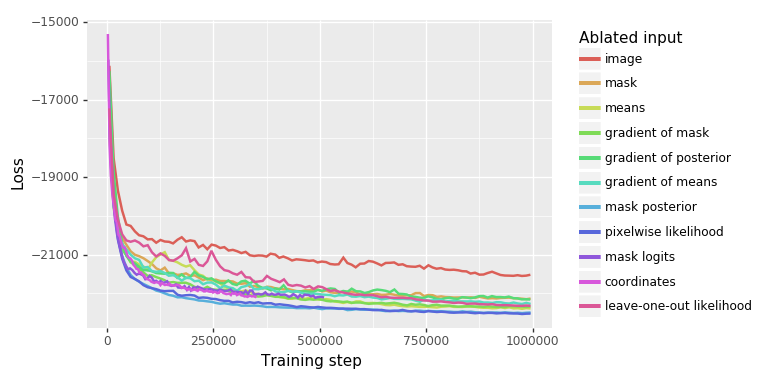}
    \caption{Ablation study for the model's total loss on Tetris. Each curve denotes the result of training the model without a particular input.}
    \vspace*{-5pt}
    \label{fig:abl_tetris_loss}
\end{figure*}

\begin{figure*}[t!]
    \centering
    \includegraphics[width=0.92\textwidth]{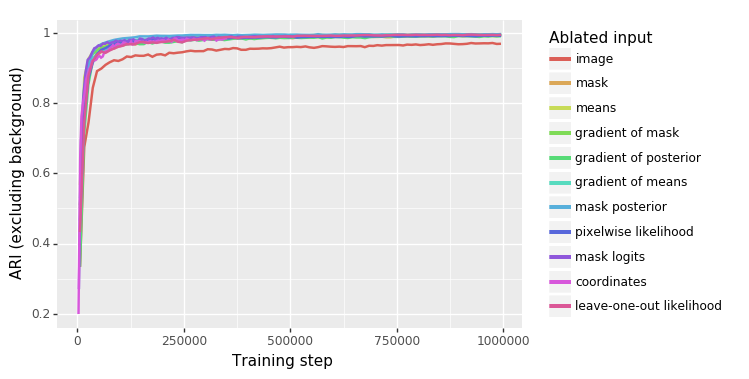}
    \caption{Ablation study for the model's segmentation performance in terms of ARI (excluding the background pixels) on Tetris. Each curve denotes the result of training the model without a particular input.}
    \vspace*{-5pt}
    \label{fig:abl_tetris_ari}
\end{figure*}

\begin{figure*}[t!]
    \centering
    \includegraphics[width=0.92\textwidth]{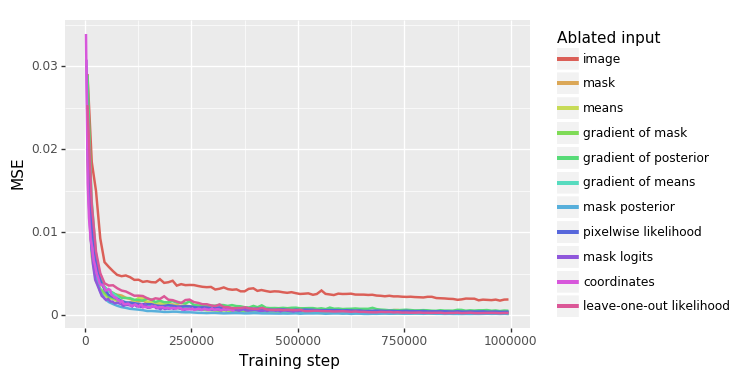}
    \caption{Ablation study for the model's reconstruction loss term on Tetris. Each curve denotes the result of training the model without a particular input. The y-axis shows the mean squared error between the target image and the output means (of the final iteration) as a proxy for the full reconstruction loss.}
    \vspace*{-5pt}
    \label{fig:abl_tetris_mse}
\end{figure*}

\begin{figure*}[t!]
    \centering
    \includegraphics[width=0.92\textwidth]{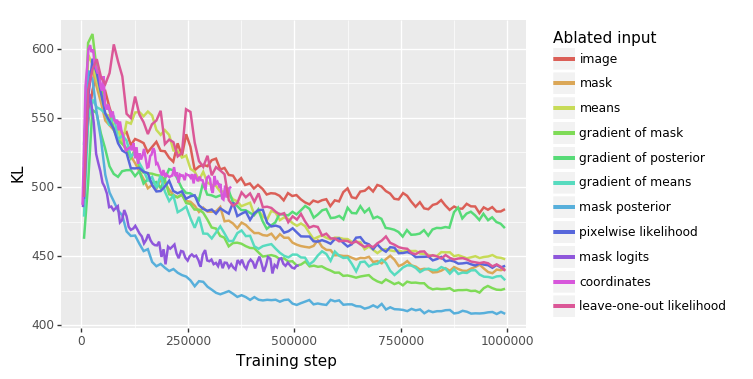}
    \caption{Ablation study for the model's KL divergence loss term on Tetris, summed over slots and iterations. Each curve denotes the result of training the model without a particular input.}
    \vspace*{-5pt}
    \label{fig:abl_tetris_kl}
\end{figure*}

\paragraph{Input Ablations}
Figures~\ref{fig:abl_clevr_loss}--\ref{fig:abl_tetris_kl} give an overview of the impact of each of the inputs to the refinement network on the total loss, mean squared reconstruction error, KL divergence loss term, and the ARI segmentation performance (excluding the background pixels) on the CLEVR and Tetris datasets.

\end{document}